\documentclass[lettersize,journal]{IEEEtran}
\usepackage{amsmath,amsfonts}
\usepackage{algorithmic}
\usepackage{algorithm}
\usepackage{array}
\usepackage{textcomp}
\usepackage{stfloats}
\usepackage{url}
\usepackage{verbatim}
\usepackage{subfloat}
\usepackage{booktabs}
\usepackage{multirow}
\usepackage{graphicx}
\usepackage{cite}
\usepackage[font=normalsize,labelfont=normalsize,caption=false,format=plain]{subfig}
\usepackage[table]{xcolor}
\usepackage[colorlinks=true,
citecolor=green!70!black,
linkcolor=red,
anchorcolor=black,
urlcolor=magenta]{hyperref}
\usepackage{colortbl}
\usepackage{placeins}
\usepackage{cleveref}
\usepackage{amssymb}

\newcommand{\myrule}{\specialrule{1pt}{.1pt}{.1pt}}
\newcommand{\mytinyrule}{\specialrule{.3pt}{.1pt}{.1pt}}

\DeclareMathOperator*{\argmin}{arg\,min}

\definecolor{b-reconstruction}{HTML}{ffffff}
\definecolor{b-embedding}{HTML}{ffffff}
\definecolor{b-flow}{HTML}{ffffff}

\definecolor{reconstruction}{HTML}{ffffff}
\definecolor{embedding}{HTML}{ffffff}
\definecolor{flow}{HTML}{ffffff}
\definecolor{separate}{HTML}{C0C0C0}

\begin{document}

\title{VQ-Flow: Taming Normalizing Flows for Multi-Class Anomaly Detection via Hierarchical Vector Quantization}

\author{Yixuan Zhou, Xing Xu, Zhe Sun, Jingkuan Song, Andrzej Cichocki~\IEEEmembership{Fellow,~IEEE}, Heng Tao Shen~\IEEEmembership{Fellow,~IEEE}
% \thanks{This work was sponsored in part by the National Natural Science Foundation of China under Grants (No. 62072080 and No. 62222203) and the New Cornerstone Science Foundation through the XPLORER PRIZE.}

\thanks{This work was supported by the National Natural Science Foundation of China under Grant 62222203 and  the New Cornerstone Science Foundation through the XPLORER PRIZE.}

\thanks{Y. Zhou, X. Xu, J. Song, H.T. Shen are with the Center for Future Media and School of Computer Science and Engineering, University of Electronic Science and Technology of China, Chengdu 611731, China (E-mail: yxzhou@std.uestc.edu.cn; xing.xu@uestc.edu.cn; jingkuan.song@gmail.com; shenhengtao@hotmail.com).

X. Xu, J. Song, H.T. Shen are also with the College of Electronic and Information Engineering, Tongji University, Shanghai 201804, China.

Zhe Sun is with Juntendo University, Tokyo 113-0033, Japan. (E-mail: z.sun.kc@juntendo.ac.jp).

Andrej Cichocki is with System Research Institute of Polish Academy of Sciences, Warsaw 01-447, Poland, and Tensor Learning Lab, RIKEN AIP, Tokyo 103-0027, Japan.
(E-mail: A.Cichocki@skoltech.ru).
}% <-this % stops a space
}

% The paper headers
\markboth{IEEE Transactions on Multimedia,~Vol.~XX, No.~XX, June~2024}%
{Shell \MakeLowercase{\textit{et al.}}: A Sample Article Using IEEEtran.cls for IEEE Journals}

\maketitle

\begin{abstract}
     Normalizing flows, a category of probabilistic models famed for their capabilities in modeling complex data distributions, have exhibited remarkable efficacy in unsupervised anomaly detection. This paper explores the potential of normalizing flows in multi-class anomaly detection, wherein the normal data is compounded with multiple classes without providing class labels. Through the integration of vector quantization (VQ), we empower the flow models to distinguish different concepts of multi-class normal data in an unsupervised manner, resulting in a novel flow-based unified method, named VQ-Flow. Specifically, our VQ-Flow leverages hierarchical vector quantization to estimate two relative codebooks: a Conceptual Prototype Codebook (CPC) for concept distinction and its concomitant Concept-Specific Pattern Codebook (CSPC) to capture concept-specific normal patterns. The flow models in VQ-Flow are conditioned on the concept-specific patterns captured in CSPC, capable of modeling specific normal patterns associated with different concepts. Moreover, CPC further enables our VQ-Flow for concept-aware distribution modeling, faithfully mimicking the intricate multi-class normal distribution through a mixed Gaussian distribution reparametrized on the conceptual prototypes. Through the introduction of vector quantization, the proposed VQ-Flow advances the state-of-the-art in multi-class anomaly detection within a unified training scheme, yielding the Det./Loc. AUROC of 99.5\%/98.3\% on MVTec AD. The codebase is publicly available at \url{https://github.com/cool-xuan/vqflow}.
\end{abstract}


\section{Introduction}

\label{sec:intro}
Anomaly detection, a fundamental task in computer vision, has been widely applied in various real-world applications, such as industrial quality control~\cite{dataset:visa, dataset:mvtec}, medical diagnosis~\cite{app:medical}, and surveillance systems~\cite{app:video, zyx:bnwvad,tmm:wvad,tmm:contrastive}.
Due to the scarcity of anomalies and high costs of abnormal data collection, anomaly detection is often formulated as an unsupervised learning problem, where the model is trained to estimate normal data distribution and detect anomalies as outliers.
Under the ideal assumption that the normal data is homogeneous, existing unsupervised anomaly detection (UAD) methods~\cite{scad:draem,scad:fcdd,scad:panda,scad:patchcore,scad:msflow} are mainly designed for single-class anomaly detection (SCAD), where the model is solely trained on normal data belonging to a particular class.
However, coming to real-world scenarios, the normal data is often composed of multiple distinct classes~\cite{mcad:uniad}, and the anomalies can be present in all classes. Take industrial quality control as an example, defects may occur on all types of products~\cite{dataset:mvtec, scad:msflow}.
When trained on the normal data mixed with multiple classes, previous SCAD methods are confused by the mixed normal distribution and degrade in performance.

The practical demand for multi-class anomaly detection (MCAD) has driven the emergence of training a unified model on the normal data blended with distinct classes and detecting anomalies across all classes in.
Most existing MCAD approaches primarily hinge on reconstruction errors, wherein models are trained to reconstruct input data and subsequently identify anomalies by evaluating the reconstruction errors~\cite{mcad:uniad, mcad:hvqTrans, mcad:omnial}.

\begin{figure*}[t]
  \centering
  \includegraphics[width=\textwidth]{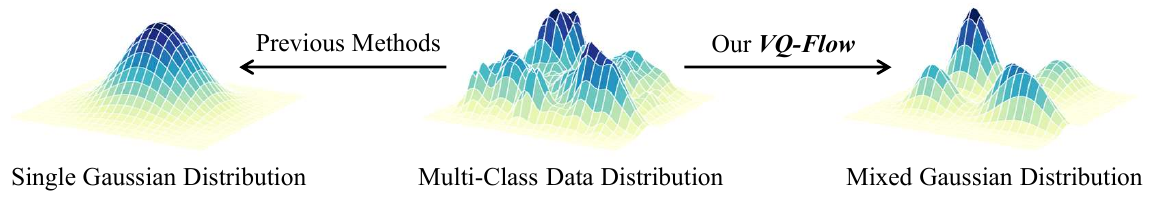}
  \caption{Previous unified methods are prone to model the intricate multi-class data distribution to a simplistic single Gaussian distribution, while our VQ-Flow transforms to the mixed Gaussian distribution through concept-aware distribution modeling.}\label{fig:problem}
  % \vspace{-8pt}
\end{figure*}

More recently, following the embedding-based paradigm~\cite{scad:patchcore}, some methods~\cite{mcad:uniformaly} leverage the memory bank to accommodate feature prototypes of multi-class normal data, detecting anomalies with large divergence from the prototypes.
Despite advancements, the essential challenge of MCAD remains unsolved, as the models are constrained to estimate all class-specific normal patterns in a unified manner, whose distribution is intricate and entangled across different classes. Without any class priors, previous unified approaches are prone to model the complicated multi-class normal data distribution to a simplistic single Gaussian distribution as shown in the left part of Fig.~\ref{fig:problem}
, leading to suboptimal performance.

\begin{figure*}[t]
  \centering
  \includegraphics[width=0.95\textwidth]{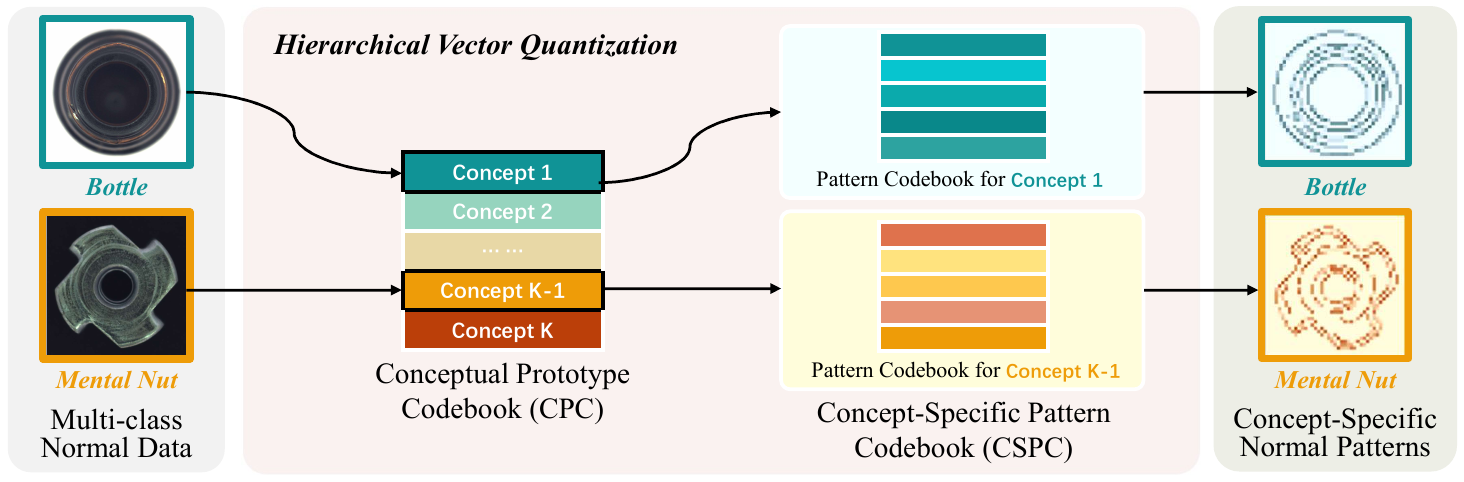}
  \caption{The illustration of our main idea in VQ-Flow for multi-class anomaly detection. Multi-class normal data is quantized through hierarchical vector quantization into a Conceptual Prototype Codebook for concept distinction and a Concept-Specific Pattern Codebook to capture concept-specific normal patterns. \emph{Best viewed in color.}
  } \label{fig:intro}
  % \vspace{-8pt}
\end{figure*}

In this study, we pivot our attention towards normalizing flows, which are acclaimed for their capabilities in characterizing complex data distributions~\cite{nf:nice, nf:realnvp, nf:glow} and performance superiority in SCAD~\cite{scad:padim, scad:cflow, scad:msflow}, to investigate their potential in multi-class anomaly detection.
To equip normalizing flows with the concept distinction without class labels, we integrate vector quantization into the flow models and propose a novel unified method VQ-Flow.
Explicated in \Cref{fig:intro}, to separate different inner classes of normal data, vector quantization (VQ) is employed to infer conceptual prototypes that represent discrete concepts within the multi-class normal data, all of which are cataloged in the Conceptual Prototype Codebook (CPC).
For each distinctive concept grasped in CPC, our VQ-Flow further adaptively captures its specific normal patterns into the Concept-Specific Pattern Codebook (CSPC) through hierarchical vector quantization as illustrated in \Cref{fig:intro}.
Conditioned on these concept-specific patterns from the CSPC, the flow models in VQ-Flow are adept at modeling specific normal patterns associated with different concepts.
Furthermore, the conceptual prototypes in CPC enable our VQ-Flow for concept-aware distribution modeling, mimicking the intricate multi-class data distribution via a mixed Gaussian distribution that is reparametrized using the conceptual prototypes, as presented in the right part of \Cref{fig:problem}.

We evaluate the proposed VQ-Flow on two industrial anomaly detection benchmarks including MVTec AD~\cite{dataset:mvtec} and VisA~\cite{dataset:visa} as well as the semantic anomaly detection dataset CIFAR-10~\cite{dataset:cifar10}. By deploying vector quantization for concept-specific normal pattern capturing and concept-aware distribution modeling, VQ-Flow advances state-of-the-art (SOTA) in multi-class anomaly detection, yielding an impressive AUROC of 99.5\% for detection and  98.3\% for localization on the MVTec AD benchmark.
On CIFAR-10, VQ-Flow also surpasses previous unified methods~\cite{mcad:uniad, mcad:omnial}.
Our main contributions are as follows:
\begin{itemize}
  \item We pioneer the exploration of normalizing flows for multi-class anomaly detection, augmenting them with vector quantization to introduce a novel unified method termed VQ-Flow.
  \item VQ-Flow leverages hierarchical vector quantization to estimate conceptual prototypes within CPC for discerning concepts and to capture concept-specific normal patterns in CSPC, thereby enabling the flow models to discriminate inner classes and master their respective normal patterns.
  \item Our VQ-Flow performs concept-aware distribution modeling via reparametrization upon the conceptual prototypes, facilitating the fidelity of the complicated multi-class data distribution.
\end{itemize}

\section{Related Work}

In this section, we first review the overall landscape of anomaly detection and then delve into the existing methods for multi-class anomaly detection.
And then, we provide a brief overview of previous approaches of utilizing normalizing flows for single-class anomaly detection.
We also introduce vector quantization and its applications in other fields, mainly in generative models and one recent method for multi-class anomaly detection, which is closely related to our VQ-Flow.

\subsubsection{Anomaly Detection}
Anomaly detection, a fundamental task across various domains such as industrial quality control~\cite{dataset:mvtec, dataset:visa}, medical diagnosis~\cite{app:medical}, surveillance systems~\cite{app:video, zyx:bnwvad,tmm:wvad,tmm:contrastive}, and financial fraud detection~\cite{ngai2011application}, aims to identify data instances that deviate significantly from the normal data distribution.
In the field of multimedia, anomaly detection has been widely applied to video surveillance~\cite{zyx:bnwvad}, image quality assessment~\cite{scad:msflow}. 
\cite{XD} demonstrates the importance of audio information in video anomaly detection, and first propose a video anomaly dataset named XD-Violence, where both video and audio data are available.
Many application-driven approaches~\cite{scad:cflow,scad:msflow,scad:fcdd,scad:patchcore,scad:simplenet} focusing on industrial defect detection~\cite{dataset:mvtec, dataset:visa} also leverage anomaly detection techniques~\cite{ruff2018deep} to ensure product quality.

\subsubsection{Multi-Class Anomaly Detection}

The challenge of multi-class anomaly detection (MCAD) arises due to the intermingling of patterns across different categories, making it a complex area in anomaly detection research. Traditionally, most MCAD techniques have relied on reconstruction errors as a key mechanism for identifying anomalies, a method detailed in several studies (e.g., \cite{mcad:uniad, mcad:omnial,tmm:ad}). In recent times, there has been a shift towards leveraging more advanced neural architectures like transformers and diffusion models~\cite{mcad:vitad, mcad:hvqTrans, mcad:diad} to improve detection capabilities. Additionally, some approaches have experimented with embedding-based methods, incorporating a memory bank for storing representations to aid anomaly detection~\cite{mcad:uniformaly, scad:patchcore}.
Despite these advancements, the exploration into the use of normalizing flows in MCAD has been overlooked. This is notable because normalizing flows have demonstrated significant potential in single-class anomaly detection (SCAD) scenarios, as seen in various research~\cite{scad:padim, scad:cflow, scad:msflow}. Given their ability to model complex data distributions effectively, investigating their application in MCAD could provide a promising avenue for future research.

\subsubsection{Flow-based Anomaly Detection}

Normalizing flows, recognized for their prowess in generative modeling~\cite{nf:realnvp, nf:glow}, have also gained traction in the domain of unsupervised anomaly detection~\cite{scad:padim,scad:cflow, scad:msflow}. These models are adept at capturing the intricacies of data distributions, a trait that particularly benefits anomaly detection tasks. Their fundamental strength lies in their invertible nature, which allows for an explicit characterization of the probability distribution of normal data while inherently associating lower likelihoods with anomalous data points.

Inspired by the effectiveness of flow-based methods in the realm of single-class anomaly detection~\cite{scad:cflow, scad:msflow}, this study delves into the viability of employing normalizing flows for multi-class anomaly detection (MCAD). To this end, we introduce a novel technique, VQ-Flow, which leverages vector quantization to confront the issue of overlapping patterns among various classes within MCAD, providing a more distinct delineation between normal and abnormal classes.

\subsubsection{Vector Quantization}

Driven by the balance between the fidelity of representation and computational efficiency, vector quantization (VQ) is already-intensely-studied~\cite{vq:stackedQuantizers, vq:survey} and widely applied to various fields, such as image compression~\cite{vq:zxs}, generative modeling~\cite{vq:vqgan, vq:vqvae}. In particular, HVQ-Trans~\cite{mcad:hvqTrans}, a reconstruction-based method for MCAD, also utilizes hierarchical vector quantization into MCAD, seeming to have an inspiration collision with our VQ-Flow.
However, the VQ incorporation in HVQ-Trans is only proposed to quantize hierarchical, or multi-stage features, while we leverage VQ~\cite{vq:stackedQuantizers} to hierarchically estimate the conceptual prototypes and their concept-specific normal patterns, enabling our VQ-Flow with concept distinction in an unsupervised manner.

% \subsubsection{Vector Quantization}

% Driven by the balance between the fidelity of representation and computational efficiency, vector quantization (VQ) is already-intensely-studied~\cite{vq:stackedQuantizers, vq:survey} and widely applied to various fields, such as image compression~\cite{vq:zxs}, generative modeling~\cite{vq:vqgan, vq:vqvae}. In particular, HVQ-Trans~\cite{mcad:hvqTrans}, a reconstruction-based method for MCAD, also utilizes hierarchical vector quantization into MCAD, seeming to have an inspiration collision with our VQ-Flow.
% However, the VQ incorporation in HVQ-Trans is only proposed to quantize hierarchical, or multi-stage features, while we leverage VQ~\cite{vq:stackedQuantizers} to hierarchically estimate the conceptual prototypes and their concept-specific normal patterns, enabling our VQ-Flow with concept distinction in an unsupervised manner.

\section{Preliminaries}

Before delving into the details of VQ-Flow, we first introduce the basic concepts and formulations of normalizing flows and vector quantization. 

\subsection{Normalizing Flows}

Normalizing flows~\cite{nf:nice, nf:realnvp, nf:glow} are probabilistic models that explicitly transform arbitrary complex distribution $p_\mathcal{X}(\boldsymbol{x})$ to a tractable base distribution $p_\mathcal{Z}(\boldsymbol{z})$ through the flow model $F$: $\mathcal{X} \leftrightarrow  \mathcal{Z}$.
Given a data sample $\boldsymbol{x}$, $F$ builds a bijective mapping between $\boldsymbol{x}$ and the latent variable $\boldsymbol{z}$ by a sequence of invertible transformations $\{f_1, f_2, ..., f_K\}$ as follows:
\begin{align}
  \boldsymbol{z} &= F(\boldsymbol{x}) \quad = f_K \circ f_{K-1} \circ ... \circ f_1(\boldsymbol{x}), \\
  \boldsymbol{x} &= F^{-1}(\boldsymbol{z}) = f_1^{-1} \circ f_2^{-1} \circ ... \circ f_K^{-1}(\boldsymbol{z}).
\end{align}
where $f_k$ is a flow block parameterized by neural networks and $F^{-1}$ is the inverse transformation of $F$.
Benefiting from the invertibility of the flow model $F$, the probability density function (PDF) of $\boldsymbol{x}$ can be computed by the change of variables as follows:
\begin{equation}
  p_\mathcal{X}(\boldsymbol{x}) = p_\mathcal{Z}(F(\boldsymbol{x})) \left| \det \frac{\partial F}{\partial \boldsymbol{x}} \right|,
\end{equation}
where $\left|\textrm{det}\frac{\partial F}{\partial \boldsymbol{x}}\right|$ is the determinant of the Jacobian matrix of $F$.
The flow model $F$ can be trained with maximum likelihood estimation (MLE) by minimizing the negative log-likelihood (NLL) of the transformed distribution $p_\mathcal{Z}(F(\boldsymbol{x}))$ with respect to the base distribution $\mathcal{Z}$ as follows:
\begin{align} \label{eq:loss_f1}
  \mathcal{L}_f 
  & = -\mathbb{E}_{\boldsymbol{x} \sim p_\mathcal{X}} \left[ \log p_\mathcal{Z}(F(\boldsymbol{x})) \right] \nonumber \\
  & =-\mathbb{E}_{\boldsymbol{x} \sim p_\mathcal{X}} \left[ \log p_\mathcal{Z}(\boldsymbol{z}) + \log \left| \det \frac{\partial F}{\partial \boldsymbol{x}} \right| \right],
\end{align}
Thanks to the invertible, in other words, lossless transformation of normalizing flows, the flow models explicitly capture the likelihood of the normal distribution $p_\mathcal{X}(\boldsymbol{x}^+)$ and detect anomalies $\boldsymbol{x}^-$ as outliers with low likelihood, yielding superior performance in unsupervised anomaly detection~\cite{scad:padim, scad:cflow, scad:msflow}.

\subsection{Vector Quantization}

Vector Quantization (VQ) aims to represent a high-dimensional space $\mathcal{X} \subset \mathbb{R}^d$ with a finite set termed codebook $\boldsymbol{C} = \{\boldsymbol{c}_1, \boldsymbol{c}_2, \ldots, \boldsymbol{c}_K\} \subset \mathbb{R}^d$, through a quantization mapping $Q: \mathcal{X}  \rightarrow C$.
Given an input vector $\boldsymbol{x}$, the quantization mapping $Q$ assigns $\boldsymbol{x}$ to $\hat{\boldsymbol{x}} = \boldsymbol{c}_{k^*} = Q(\boldsymbol{x})$, where $k^*$ is the index of the most similar codeword to $\boldsymbol{x}$.
To approximate the input distribution $p_\mathcal{X}(\boldsymbol{x})$ with the quantized distribution $p_C(\hat{\boldsymbol{x}})$, the loss objective $\mathcal{L}_Q$ is often defined as a mean squared error (MSE) loss as follows:
\begin{equation}
  \mathcal{L}_Q = \mathbb{E}[\|\boldsymbol{x} - \hat{\boldsymbol{x}}\|^2] \quad \textrm{and} \quad \hat{\boldsymbol{x}} = Q(\boldsymbol{x}).
\end{equation}
Vector quantization provides a balance between the fidelity of representation and computational efficiency, driving its applications in various fields~\cite{vq:zxs,vq:vqgan,vq:stackedQuantizers}.
In our VQ-Flow, we share the same spirit with VQGAN~\cite{vq:vqgan} in utilizing vector quantization to disentangle conceptual prototypes that capture the high-level semantics of the input data. This insight is the cornerstone of our VQ-Flow to tackle multi-class anomaly detection.

\section{Method}

In this section, we first revisit flow-based methods proposed for anomaly detection under the single-class scenario and investigate the challenges of extending them to multi-class settings in Section~\ref{subsec:revisit}. We then elaborate on the details of our VQ-Flow for the extension of flow-based methods to multi-class scenarios.

\subsection{Revisiting Flow-based Single-Class Anomaly Detection}
\label{subsec:revisit}

Rather than directly modeling $p_\mathcal{X}(\boldsymbol{x})$, most previous flow-based methods~\cite{scad:padim,scad:msflow,scad:cflow} for SCAD utilizes normalizing flows to model the multi-scale representative features $\{\boldsymbol{h}_i \in \mathbb{R}^{D_i \times H_i \times W_i}\}_{i=1}^L=E(\boldsymbol{x})$ extracted from the raw data $\boldsymbol{x}$ with a pre-trained feature extractor $E$.
Under the single-class scenario, the flow models only need to overfit certain normal patterns within one class and yield impressive performance.
Beyond overfitting to a particular type of normal patterns, multi-class anomaly detection (MCAD)~\cite{mcad:uniad} requires a model to learn and differentiate between the varied normal patterns of multiple classes within a single unified model. 
When directly adopting the flow-based SOTA~\cite{scad:msflow} to the multi-class scenario, the performance degrades significantly, as the flow models struggle to model the entangled normal patterns across different classes.
Therefore, the main challenge of extending flow-based methods to MCAD lies in enabling the flow models to be semantically aware without explicit class annotations, and further capable of capturing the specific normal patterns associated with each class.

\begin{figure*}[t]
  \centering
  \includegraphics[width=\textwidth]{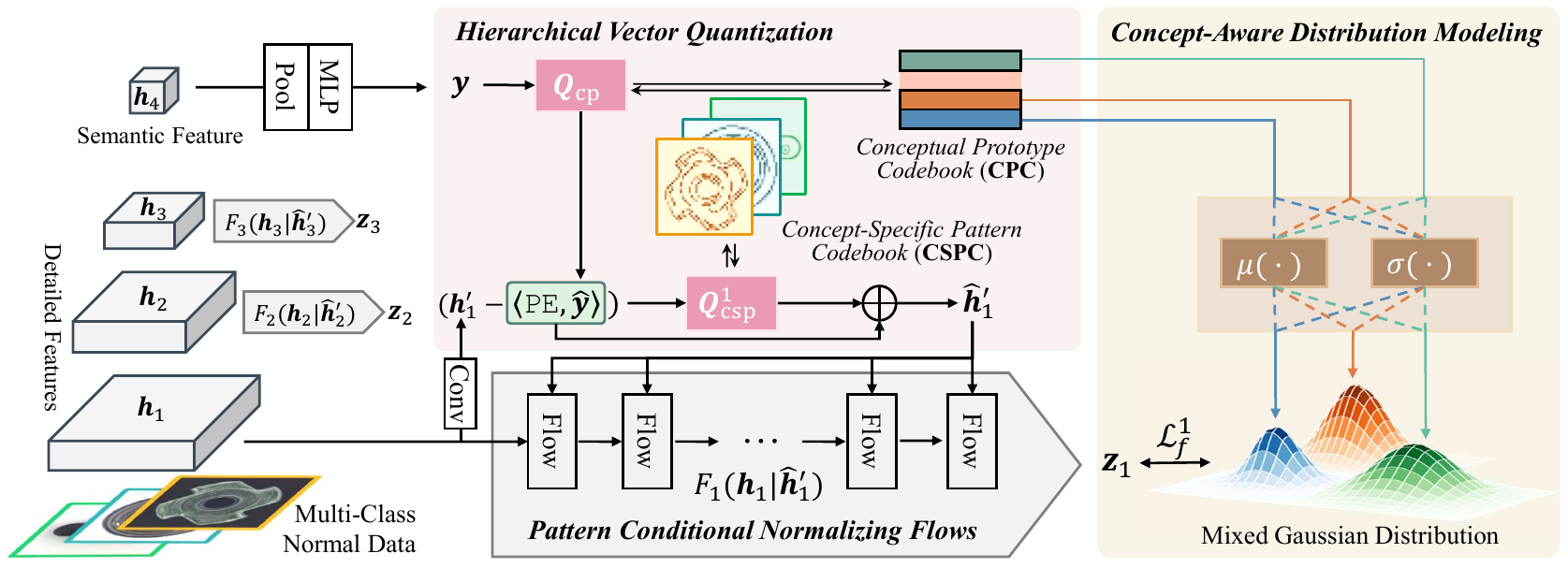}
  \caption{The overview of our proposed VQ-Flow for multi-class anomaly detection.
  For clarity, only the hierarchical vector quantization for the first branch of flow models $F_1$ is depicted, and the other flow models $F_2$ and $F_3$ are overlooked.
  % \emph{Best viewed in color.}
  }\label{fig:framework}
  % \vspace{-8pt}
\end{figure*}

\subsection{Conceptual Prototype Codebook for Concept Distinction}
\label{subsec:ccp}

In MCAD, the models only need to grasp different concepts within the inner classes of multi-class normal data, without the need to match the actual semantic classes.
Therefore, our VQ-Flow first utilizes vector quantization, which shares the same spirit with \textit{k}-means clustering, to estimate the conceptual prototypes that are distinctive from each other and represent a specific concept within the inner classes of MCAD.
Specifically, we perform vector quantization upon the high-level features $\boldsymbol{h}_L$ (e.g., $\boldsymbol{h}_4$, the features of stage 4 in ResNet~\cite{backbone:resnet}) with semantic perception.
For computational efficiency, $\boldsymbol{h}_L$ are first aggregated into a single feature vector by average pooling and then projected into a compressed feature space $\boldsymbol{y} \in \mathbb{R}^{D_{cp}}$ by an MLP layer as follows:
\begin{equation}
  \boldsymbol{y} = \textrm{MLP}(\textrm{AvgPool}(\boldsymbol{h}_L)),
\end{equation}
where $D_{cp}$ also refers to the dimension of the conceptual prototypes.
The semantic-sensitive feature vectors $\boldsymbol{y} \in \mathbb{R}^{D_{cp}}$ are then quantized into the Conceptual Prototype Codebook (CPC) $\boldsymbol{C}_\textrm{cp}$ with $K_\textrm{cp}$ prototypes, through the quantization mapping $Q_\textrm{cp}$ as follows:
\begin{equation}
  \hat{\boldsymbol{y}} = Q_\textrm{cp}(\boldsymbol{y}) = \argmin_{\boldsymbol{c}_k \in \boldsymbol{C}_\textrm{cp}} \|\boldsymbol{y} - \boldsymbol{c}_k\|^2,
\end{equation}
where $\hat{\boldsymbol{y}}$ is the most similar prototype to $\boldsymbol{y}$ in CPC.
Correspondingly, the training objective for CPC learning is formulated as follows:
\begin{equation} \label{eq:loss_Qcp}
  \mathcal{L}_{Q_\textrm{cp}} = \mathbb{E}[\|\boldsymbol{y} - \hat{\boldsymbol{y}}\|^2].
\end{equation}
For intuitive understanding, the conceptual prototypes $\hat{\boldsymbol{y}}$ in codebook $\boldsymbol{C}_\textrm{cp}$ can be viewed as 
pseudo class labels of inner classes within MCAD as depicted in \Cref{fig:intro}.
Despite the uncertainty of matching the real semantic classes, each quantized conceptual prototype in CPC is distinctive from others, endowing our VQ-Flow with concept discrimination.

\begin{figure*}[t]
  \centering
  \includegraphics[width=0.8\textwidth]{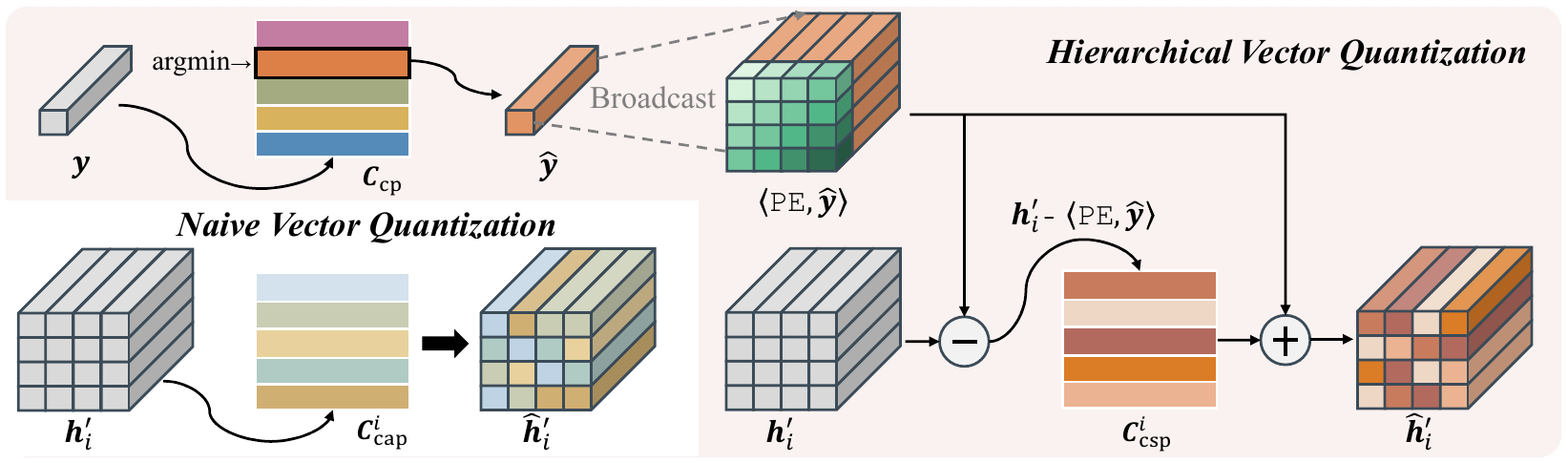}
  \caption{The illustration of hierarchical (residual) vector quantization for Conceptual Prototype Codebook $\boldsymbol{C}_\textrm{cp}$ and Concept-Specific Pattern Codebook $\boldsymbol{C}_\textrm{csp}$ learning in our VQ-Flow. 
  The naive quantization for Concept-Agnostic Pattern Codebook $\boldsymbol{C}_\textrm{cap}$ is also depicted for comparison.}
  % \vspace{-8pt}
  \label{fig:quantization}
\end{figure*}

\subsection{Concept-Specific Pattern Codebook for Pattern Capturing}
\label{subsec:cspc}

In addition to concept distinction, our VQ-Flow also captures the concept-specific normal patterns, which is significant for industrial anomaly detection where the normal layout patterns vary across different classes of products.
Similar to the dimension reduction of the conceptual prototypes, multi-scale feature maps $\{\boldsymbol{h}_i\}_{i=1}^3$ with detailed spatial information are first projected to $\{\boldsymbol{h}^{\prime}_i \in \mathbb{R}^{D^{\prime}_i \times H_i \times W_i}\}_{i=1}^3$ within a lower-dimensional feature space.

\subsubsection{Naive Quantization Formulation}

The straightforward approach of capturing the normal patterns is directly quantizing $\boldsymbol{h}^{\prime}_i$ through the pixel-wise quantization mapping $Q_\textrm{cap}^i$ as follows:
\begin{equation}
  \hat{\boldsymbol{h}^{\prime}_i}[m,n] = Q_\textrm{cap}^i(\boldsymbol{h}^{\prime}_i[m,n]) = \argmin_{\boldsymbol{c}_k \in \boldsymbol{C}_\textrm{cap}^i} \|\boldsymbol{h}^{\prime}_i[m,n] - \boldsymbol{c}_k\|^2,
\end{equation}
where $\boldsymbol{h}^{\prime}_i[m,n]$ refers to the feature vector at $(m,n)$-th position of $\boldsymbol{h}^{\prime}_i$.
However, the naive quantization $Q_\textrm{cap}^i$ lacks the semantic perception, without any interaction with the conceptual prototypes in CPC, thus named Concept-Agnostic Pattern Codebook (CAPC).

\subsubsection{Residual Quantization Formulation}

To capture the concept-specific normal patterns, the proposed VQ-Flow leverage the residual quantization~\cite{vq:zxs, vq:stackedQuantizers} to hierarchically quantize detailed feature maps $\{\boldsymbol{h}^{\prime}_i\}_{i=1}^3$ conditioned on the conceptual prototypes in CPC, and estimate the Concept-Specific Pattern Codebook (CSPC), as depicted in \Cref{fig:quantization}.
Specifically, we instead quantize the residual between $\boldsymbol{h}^{\prime}_i[m,n]$ and the conceptual prototype $\hat{\boldsymbol{y}}$ through $Q_\textrm{csp}^i$ as follows:
\begin{align} \label{eq:hi}
  \hat{\boldsymbol{h}^{\prime}_i}[m,n] - \hat{\boldsymbol{y}} 
  & = Q_\textrm{csp}^i(\boldsymbol{h}^{\prime}_i[m,n] - \hat{\boldsymbol{y}}) \nonumber \\
  & = \argmin_{\boldsymbol{c}_k \in \boldsymbol{C}_\textrm{csp}^i} \|(\boldsymbol{h}^{\prime}_i[m,n] - \hat{\boldsymbol{y}}) - \boldsymbol{c}_k\|^2.
\end{align}
For simplicity and clarity, we provide the feature-map level quantization formulation by viewing $\boldsymbol{h}^{\prime}_i$ as a whole, resulting in the $Q_\textrm{csp}^i$ transformed as follows:
\begin{equation}
  \hat{\boldsymbol{h}^{\prime}_i} - \hat{\boldsymbol{y}} = Q_\textrm{csp}^i(\boldsymbol{h}^{\prime}_i - \hat{\boldsymbol{y}}) = Q_\textrm{csp}^i(\boldsymbol{h}^{\prime}_i - Q_{cp}(\boldsymbol{y})),
\end{equation}
where $\hat{\boldsymbol{y}}$ is broadcasted across the entire feature map $\boldsymbol{h}^{\prime}_i$, and $Q_\textrm{csp}^i(\boldsymbol{h}^{\prime}_i - Q_{cp}(\boldsymbol{y}))$ can be reformulated as $Q_\textrm{csp}^i(\boldsymbol{h}^{\prime}_i | Q_{cp}(\boldsymbol{y}))$ to explicitly demonstrate the hierarchical quantization of CSPC conditioned on CPC.
Despite the residual form, the training objective of CSPC follows the standard quantization loss:
\begin{equation} \label{eq:loss_Qcsp}
  \mathcal{L}_{Q_\textrm{csp}^i} = \mathbb{E}[\|\boldsymbol{h}^{\prime}_i - \hat{\boldsymbol{h}^{\prime}_i}\|^2].
\end{equation}
The quantized feature map $\hat{\boldsymbol{h}^{\prime}_i}$ is computed by adding the quantized residual $Q_\textrm{csp}^i(\boldsymbol{h}^{\prime}_i - Q_{cp}(\boldsymbol{y}))$ back to the conceptual prototype $\hat{\boldsymbol{y}}$ as follows:
\begin{equation}
  \hat{\boldsymbol{h}^{\prime}_i} = Q_\textrm{csp}^i(\boldsymbol{h}^{\prime}_i - \hat{\boldsymbol{y}}) + \hat{\boldsymbol{y}}.
\end{equation}

To additionally equip the CSPC with positional awareness that is crucial for industrial anomaly detection~\cite{dataset:mvtec,dataset:visa}, we further concatenate the positional embedding $\texttt{PE}$ to with $\hat{\boldsymbol{y}}$ before residual quantization, resulting in
% With the incorporation of $\texttt{PE}$, the quantized feature map $\hat{\boldsymbol{h}^{\prime}_i}$ is computed as follows:
\begin{equation}
  \hat{\boldsymbol{h}^{\prime}_i} = Q_\textrm{csp}^i(\boldsymbol{h}^{\prime}_i - \langle \texttt{PE}, \hat{\boldsymbol{y}} \rangle) + \langle \texttt{PE}, \hat{\boldsymbol{y}} \rangle,
\end{equation}
where $\langle \cdot \rangle$ denotes the concatenation operator. In particular, due to the request of aligning channel dimensions for residual quantization, $\{D^{\prime}_i\}_{i=1}^3$ are all set to $D_{csp}=D_{cp} + D_\texttt{PE}$, where $D_\texttt{PE}$ is the dimension of the positional embedding.

With the concept-specific normal patterns captured in CSPC, our VQ-Flow inserts the quantized feature maps $\hat{\boldsymbol{h}^{\prime}_i}$ into the flow model as the condition features, and each branch of the flow model $F_i$ is formulated as follows:
\begin{equation}
  \boldsymbol{z}_i = F_i(\boldsymbol{h}_{i} | \hat{\boldsymbol{h}^{\prime}_i}) = f_{K_i} \circ f_{K_i-1} \circ ... \circ f_{1}(\boldsymbol{h}_{i} | \hat{\boldsymbol{h}^{\prime}_i}).
\end{equation}
To this end, our VQ-Flow is conditioned on the prototype-specific normal patterns in CSPC that are hierarchically derived from the conceptual prototypes in CPC, capable of distinguishing different concepts and modeling specific normal patterns associated with certain concepts under the multi-class scenario.

\subsection{Concept-Aware Conditional Distribution Modeling}

To faithfully model the intricate multi-class data distribution, we replace the transformation target of our VQ-Flow from the base distribution $p_\mathcal{Z}(\boldsymbol{z})$, usually a simplistic standard Gaussian distribution~\cite{nf:realnvp,scad:msflow}, to a mixed conditional Gaussian distribution $p_\mathcal{Z}(\boldsymbol{z} | \hat{\boldsymbol{y}})$ reparametrized on the conceptual prototypes $\hat{\boldsymbol{y}} \in \boldsymbol{C}_\textrm{cp} = \{\boldsymbol{c}_1, \boldsymbol{c}_2, \ldots, \boldsymbol{c}_{K_\textrm{cp}}\}$.
The mixed Gaussian distribution $p_\mathcal{Z}(\boldsymbol{z} | \hat{\boldsymbol{y}})$ can be presented as the sum of $K_\textrm{cp}$ concept-specific Gaussian distributions $p_\mathcal{Z}(\boldsymbol{z} | \boldsymbol{c}_k)$ dedicated to each conceptual prototype $\boldsymbol{c}_k$, explicitly formulated as follows:
\begin{equation}
  p_\mathcal{Z}(\boldsymbol{z} | \hat{\boldsymbol{y}}) = \sum_{k=1}^{K_\textrm{cp}} \mathcal{N}(\boldsymbol{z}; \mu(\boldsymbol{c}_k), \sigma(\boldsymbol{c}_k)^2), 
\end{equation}
where $\mu(\cdot)$ and $\sigma(\cdot)$ are paired neural networks for mean and variance prediction and shared across different branches of the flow model.
Specifically, when the quantized conceptual prototype $\hat{\boldsymbol{y}} = \boldsymbol{c}_{k^*}$, our VQ-Flow adaptively regulates the transformation target to its dedicated Gaussian distribution $p_\mathcal{Z}(\boldsymbol{z} | \boldsymbol{c}_{k^*})$, where $k^*$ is the index of the most similar prototype to $\boldsymbol{y}$ in CPC.

As shown in \Cref{fig:framework}, Given the the feature map $\boldsymbol{h}_1$, the transformed variable $\boldsymbol{z}_1$ yielded by the first branch of the flow model $F_1$ is computed as follows:
\begin{equation}
  \boldsymbol{z}_1 = F_1(\boldsymbol{h}_1 | \hat{\boldsymbol{h}^{\prime}_1}) \sim \mathcal{N}(\boldsymbol{z}_1; \mu(\hat{\boldsymbol{y}}), {\sigma(\hat{\boldsymbol{y}})}^2),
\end{equation}
following the Gaussian distribution reparametrized by $\mu(\hat{\boldsymbol{y}})$ and ${\sigma(\hat{\boldsymbol{y}})}^2$.
Correspondingly, the training objective of $F_1$ is reformulated as follows:
% The equation is too long,so we split it.
\begin{align} \label{eq:loss_f}
  \mathcal{L}^1_f = -\mathbb{E}_{\boldsymbol{h}_1 \sim p_\mathcal{H}} \left[ \log p_\mathcal{Z}(\boldsymbol{z}_1 | \hat{\boldsymbol{y}}) + \log \left| \det \frac{\partial F}{\partial \boldsymbol{h}_1} \right| \right] , 
\end{align}
For Gaussian distribution:
\begin{align} \label{eq:log_gauss}
  \log p_\mathcal{Z}(\boldsymbol{z}_1 | \hat{\boldsymbol{y}}) = -\frac{1}{2} \log(2\pi) - \log \sigma(\hat{\boldsymbol{y}}) - \frac{(\boldsymbol{z}_1 - \mu(\hat{\boldsymbol{y}}))^2}{2{\sigma(\hat{\boldsymbol{y}})}^2} , 
\end{align}

\subsection{Unified Training Scheme}

To train the VQ-Flow in an end-to-end manner, we provide a unified training scheme that jointly optimizes the flow models $F$ and the hierarchical vector quantization codebooks.
The overall training objective is formulated by aggregating the loss terms of the vector quantization for CPC (Eq.~\ref{eq:loss_Qcp}) and CSPC (Eq.~\ref{eq:loss_Qcsp}), as well as the flow models (Eq.~\ref{eq:loss_f}) as follows:
\begin{equation}
  \mathcal{L} = 
  \sum_{i=1}^3 \alpha_i \mathcal{L}^i_f
  + \beta \mathcal{L}_{Q_\textrm{cp}} 
  + \sum_{i=1}^3 \gamma_{i} \mathcal{L}_{Q_\textrm{csp}^i},
\end{equation}
where $\alpha_i$, $\beta$, and $\gamma_i$ are the hyperparameters that balance loss terms, which are all set to 1 in our experiments without additional tuning.

\begin{table*}[t]
  \caption{Comparison of detection and localization AUROC (Det./Loc.) on MVTec AD~\cite{dataset:mvtec} under the \textbf{Multi-Class} setting.}
  % The reconstruction/embedding/flow-based methods are denoted with background colors of $\textcolor{b-reconstruction}{\blacksquare}$/$\textcolor{b-embedding}{\blacksquare}$/$\textcolor{b-flow}{\blacksquare}$. \emph{Best viewed in color.}}
  \label{tab:comparison-mvtec-mc}
  \centering
  \resizebox{2\columnwidth}{!}{%
   \begin{tabular}{cl|>{\columncolor{reconstruction}}p{1.6cm}<{\centering}>{\columncolor{embedding}}p{1.6cm}<{\centering}>{\columncolor{flow}}p{1.6cm}<{\centering}|>{\columncolor{reconstruction}}p{1.8cm}<{\centering}>{\columncolor{reconstruction}}p{2.0cm}<{\centering}>{\columncolor{reconstruction}}p{1.8cm}<{\centering}>{\columncolor{embedding}}p{1.8cm}<{\centering}>{\columncolor{flow}}p{1.8cm}<{\centering}}
  \myrule
  \multicolumn{2}{c|}{Category}                                                    & \multicolumn{3}{c|}{Separate training scheme
  }                                         & \multicolumn{5}{c}{Unified training scheme}                                                       \\ \mytinyrule
  \multicolumn{2}{c|}{Method}                                                               & DRAEM~\cite{scad:draem}       & PatchCore~\cite{scad:patchcore}           & MSFlow~\cite{scad:msflow}             & UniAD~\cite{mcad:uniad}       & HVQ-Trans~\cite{mcad:hvqTrans}               & DiAD~\cite{mcad:diad}            & Uniformaly~\cite{mcad:uniformaly}           & \textbf{VQ-Flow}     \\ \mytinyrule
  \multicolumn{2}{c|}{Venue}                                                          & CVPR$^\prime$21     & CVPR$^\prime$22             & TNNLS$^\prime$23            & NeurIPS$^\prime$22  & NeurIPS$^\prime$23               & AAAI$^\prime$24           & ArXiv$^\prime$23             & \textbf{Ours}        \\ \mytinyrule
  \multicolumn{1}{c|}{\multirow{10}{*}{\rotatebox{270}{Object\quad}}} & \multicolumn{1}{l|}{Bottle}     & 94.6 / 87.4 & \textbf{100} / 97.4 & \textbf{100} / 97.1 & 99.7 / 98.4  & \textbf{100} / 98.3 & 99.7 / \textbf{98.4}  & \textbf{100}  / 98.1 & \textbf{100 / 98.4}  \\
  \multicolumn{1}{c|}{}                         & \multicolumn{1}{l|}{Cable}       & 61.8 / 70.4 & 95.3 / 93.6         & 89.7 / 95.1         & 95.2 / 97.3  & 99.0 / \textbf{98.1}         & 94.8 / 96.8            & 97.3 / 96.5          & \textbf{99.1} / 97.9 \\
  \multicolumn{1}{c|}{}                         & \multicolumn{1}{l|}{Capsule}     & 70.2 / 49.2 & 96.8 / 98.0         & 97.6 / 98.7         & 86.9 / 98.5 & 95.4 / 98.8         & 89.0 / 97.1          & 98.9 / 99.0 &  \textbf{99.0 /99.3}         \\
  \multicolumn{1}{c|}{}                         & \multicolumn{1}{l|}{Hazelnut}    & 95.1 / 96.0 & 99.3 / 97.6         & 99.5 / 98.2         & 99.8 / 98.1 & \textbf{100} / 98.8         &99.5 / 98.3 & \textbf{100} / 99.1  & \textbf{100 / 99.2}  \\
  \multicolumn{1}{c|}{}                         & \multicolumn{1}{l|}{Metal Nut}   & 88.9 / 72.6 & 99.1 / 96.3         & 98.6 / 98.7         & 99.2 / 94.8  & 99.9 / 96.3       &  99.1 / 97.3          & 99.9 / 97.9          & \textbf{100 / 98.8}  \\
  \multicolumn{1}{c|}{}                         & \multicolumn{1}{l|}{Pill}        & 69.0 / 90.0   & 86.4 / 90.8       & 96.9 / 97.8         & 93.7 / 95.0& 95.8 / 97.1        &  95.7 / 95.7         & \textbf{98.3} / 97.3          & 98.2 / \textbf{98.8} \\
  \multicolumn{1}{c|}{}                         & \multicolumn{1}{l|}{Screw}       & 93.3 / 89.3 & 94.2 / 98.9         & 90.6 / 98.0         & 87.5 / 98.3& 95.6 / 98.9         &  90.7 / 97.9          & 94.8 / \textbf{99.5}          & \textbf{98.4} / 99.0   \\
  \multicolumn{1}{c|}{}                         & \multicolumn{1}{l|}{Toothbrush}  & 82.8 / 94.4 & \textbf{100} / 98.8          & 97.2 / 98.4         & 94.2 / 98.4  & 93.6 / 98.6  & 99.7 / \textbf{99.0}          & \textbf{100} / 98.9  & 98.2 / 98.8          \\
  \multicolumn{1}{c|}{}                         & \multicolumn{1}{l|}{Transistor}  & 83.9 / 73.1 & 98.9 / 92.3         & 95.2 / 88.5         & 99.8 / \textbf{97.9} & 99.7 / 97.9         &  99.8 / 95.1 & 99.3 / 96.3          & \textbf{100} / 96.3          \\
  \multicolumn{1}{c|}{}                         & \multicolumn{1}{l|}{Zipper}      & 99.1 / 96.9 & 97.1 / 95.7         & 99.5 / 98.4         & 95.8 / 96.8 & 97.9 / 97.5 & 95.1 / 96.2         & 99.6 / 97.4          & \textbf{100 / 98.9}          \\ \mytinyrule
  \multicolumn{1}{c|}{\multirow{5}{*}{\rotatebox{270}{Texture\quad}}} & \multicolumn{1}{l|}{Carpet}      & 95.9 / 95.2 & 97.0 / 98.1         & 96.7 / 98.4         & 99.8 / 98.5  & \textbf{99.9} / 98.7          & 99.4 / 98.6 & 98.4 / \textbf{99.4}          & \textbf{99.9} / 98.6          \\
  \multicolumn{1}{c|}{}                         & \multicolumn{1}{l|}{Grid}        & 98.1 / 99.0 & 91.4 / 98.4         & 99.5 / 99.0         & 98.2 / 96.5 & 97.0 / 97.0 & 98.5 / 96.6              & 98.5 / \textbf{99.4}          & \textbf{99.9} / 99.1 \\
  \multicolumn{1}{c|}{}                         & \multicolumn{1}{l|}{Leather}     & 99.9 / 98.6 & 100 / 99.2          & 99.7 / 99.1         & 100 / 98.8  & \textbf{100} / 98.8            & 99.8 / 98.8  & \textbf{100 / 99.6}  & \textbf{100} / 99.0  \\
  \multicolumn{1}{c|}{}                         & \multicolumn{1}{l|}{Tile}        & 98.3 / 98.1 & 96.0 / 90.3         & 99.3 / 96.4         & 99.3 / 91.8 & 99.2 / 92.2         & 96.8 / 92.4          & 99.5 / 96.4          & \textbf{99.9 / 97.2} \\
  \multicolumn{1}{c|}{}                         & \multicolumn{1}{l|}{Wood}        & 99.8 / 96.2 & 93.8 / 90.8         & 99.4 / 94.6         & 98.6 / 93.2 & 97.2 / 92.4         & 99.7 / 93.3          & 99.8 / 95.2          & \textbf{100 / 95.5} \\ \mytinyrule
  \multicolumn{2}{c|}{Average}                                                     & 88.7 / 87.1 & 96.4 / 95.7         & 97.3 / 97.1                              & 96.5 / 96.8 & 98.0 / 97.3         & 97.2 / 96.8            & 99.2 / 98.1          & \textbf{99.5 / 98.3} \\ \myrule
  \end{tabular}%
  }
% \vspace{-5pt}
\end{table*}

\begin{table}[t]
  \centering
  \caption{Comparison of detection and localization AUROC (Det./Loc.) on VisA~\cite{dataset:visa} under the \textbf{Multi-Class} setting.}
  \label{comparison:visa}
  \renewcommand{\arraystretch}{1.2}
  \resizebox{0.9\linewidth}{!}{%
  \begin{tabular}{c|l|l|p{2.1cm}<{\centering}}
  \myrule
          & Method           & Venue & Det. / Loc. \\ \mytinyrule
  \multirow{4}{*}{\rotatebox{270}{\centering Separated\quad}} &DRAEM\cite{scad:draem}            & CVPR$^\prime$21                    & 80.5 / 87.0                             \\
                                & JNLD\cite{scad:jnld}      & ICME$^\prime$22                    & 87.1 / 95.2                           \\
                                & SimpleNet\cite{scad:simplenet}    & CVPR$^\prime$23                    & 87.2 / 96.8                           \\
                                & MSFlow\cite{scad:msflow}        & TNNLS$^\prime$23                   & 93.1  / 98.3                           \\ \hline
  \multirow{5}{*}{\rotatebox{270}{Unified\quad}}  & UniAD\cite{mcad:uniad}         & NeurIPS$^\prime$22                 & 91.9 / 98.6                           \\
                                & OmniAL\cite{mcad:omnial}       & CVPR$^\prime$23                    & 87.8 / 96.6                           \\
                                & HVQ-Trans\cite{mcad:hvqTrans}    & NeurIPS$^\prime$23                 & 93.2  /  98.7                           \\
                                & DiAD\cite{mcad:diad}          & AAAI$^\prime$24                    & 86.8 / 96.0                           \\ \cline{2-4} 
                                &  \textbf{VQ-Flow} & \textbf{Ours}              & \textbf{95.9}  / \textbf{98.9}                  \\ \myrule
  \end{tabular}
  }
\end{table}

\section{Experiment}

We validate the effectiveness of our proposed VQ-Flow on three multi-class anomaly detection datasets and compare it with previous separate or unified training methods under the multi-class setting.

\subsection{Experimental Setup}

\subsubsection{Datasets} We evaluate the proposed VQ-Flow on two industrial anomaly detection benchmarks including MVTec AD~\cite{dataset:mvtec} and VisA~\cite{dataset:visa} with multiple types of products, and the semantic anomaly detection dataset CIFAR-10~\cite
{dataset:cifar10}. 

1) \emph{MVTec AD}~\cite{dataset:mvtec} is a widely-used benchmark for industrial anomaly detection, which contains 15 categories of industrial products, 10 of objects, and 5 of textures. Each category contains normal and defective samples, and only the normal images are provided for training.

2) \emph{VisA}~\cite{dataset:visa} is a newly-released benchmark, which collects 12 categories of products with complex layouts. Similar to MVTec AD, VisA contains both normal and defective samples for each category, and only the normal images are accessible during training.

3) \emph{CIFAR-10}~\cite{dataset:cifar10} is initially designed for image classification, which contains 10 classes of natural images. We follow the same setting as~\cite{mcad:uniad} to conduct semantic anomaly detection on CIFAR-10, where a subset of 5 classes is chosen to represent the normal data while the remaining 5 classes are treated as anomalies.

\subsubsection{Evaluation Metrics} 

We adopt the area under the receiver operating characteristic curve (AUROC) as the primary evaluation metric for anomaly detection performance. In particular, for MVTec AD and VisA, we also report the localization AUROC, which measures the performance of pixel-wise anomaly detection.

\subsubsection{Implementation Details} 

We employ ConvNeXt-B~\cite{backbone:convnext} as the feature extractor for all datasets, and the previous flow-based SOTA of SCAD, MSFlow~\cite{scad:msflow} is selected as the baseline for our VQ-Flow to integrate the hierarchical vector quantization.
The feature dimension of CPC and positional embedding $D_{cp}$ and $D_\texttt{PE}$ are set to 256 and 32.
The codebook size of CPC and CSPC, $K_\textrm{cp}$ and $K_\textrm{csp}$, are set to 32 and 512, respectively. We adopt the Adam optimizer with a learning rate of $10^{-4}$ and a batch size of 16 for all experiments and train our VQ-Flow for 100 epochs on one RTX 3090 GPU card.
Specifically, input images are resized to $384\times384$ for MVTec AD and VisA, and $112\times112$ for CIFAR-10.

\begin{figure*}[t]
  \centering
  \subfloat[Det. AUROC on MVTec AD]{\includegraphics[width=0.3\linewidth]{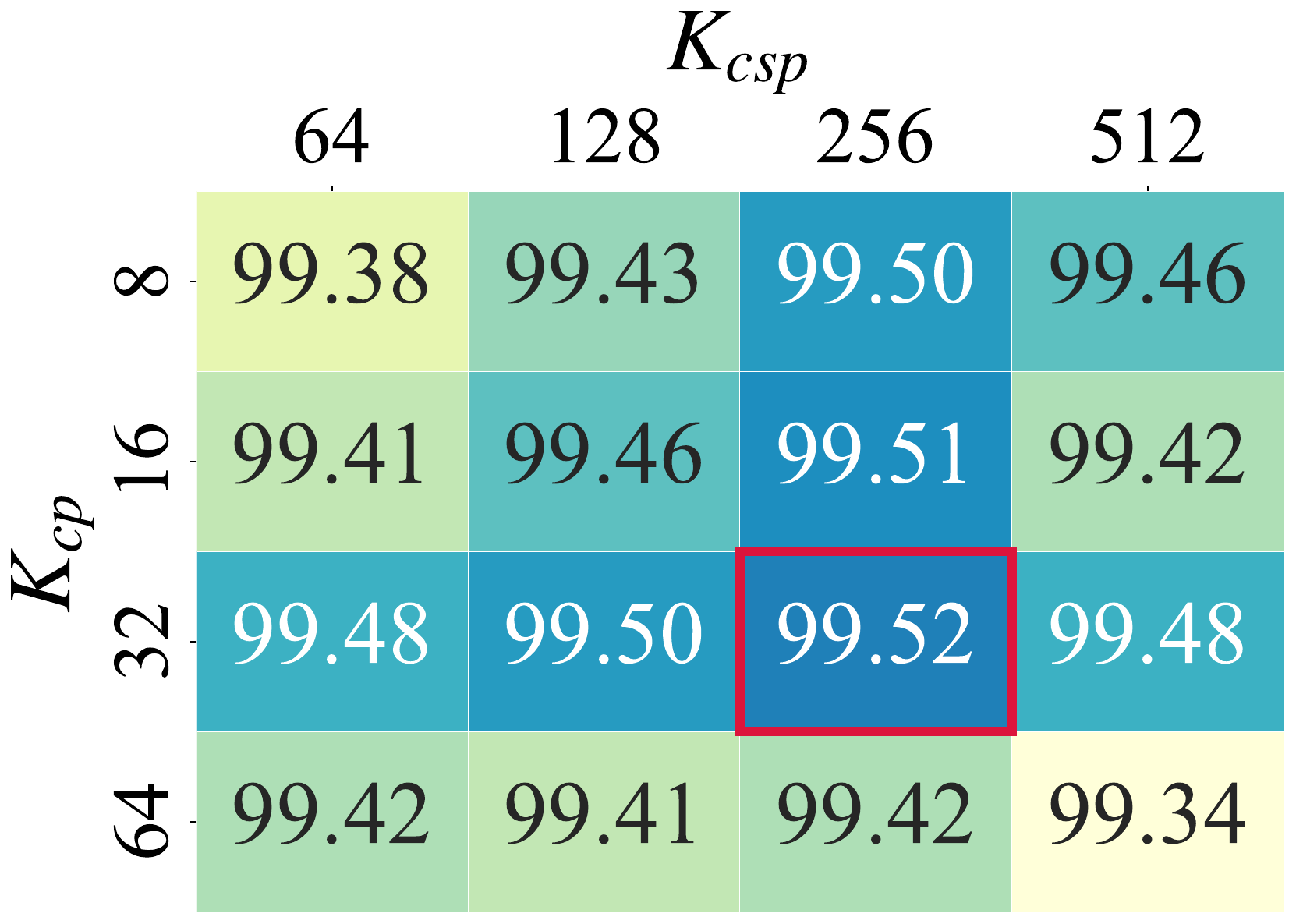}
    \label{fig:mvtec_det_heatmap}}
    \hfill
  \subfloat[Loc. AUROC on MVTec AD]{\includegraphics[width=0.3\linewidth]{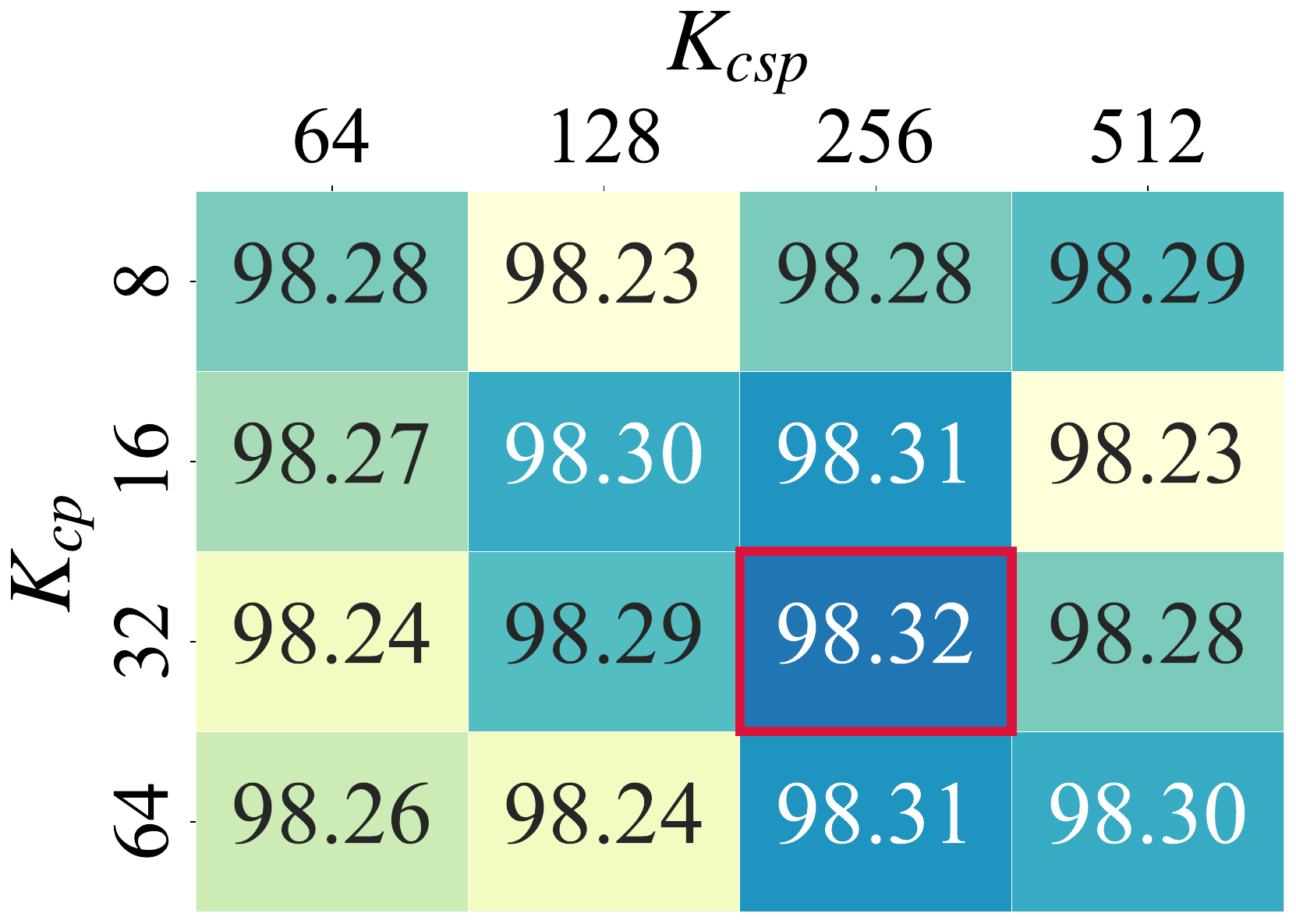}
    \label{fig:mvtec_loc_heatmap}}
    \hfill
  \subfloat[Det. AUROC on CIFAR-10]{\includegraphics[width=0.3\linewidth]{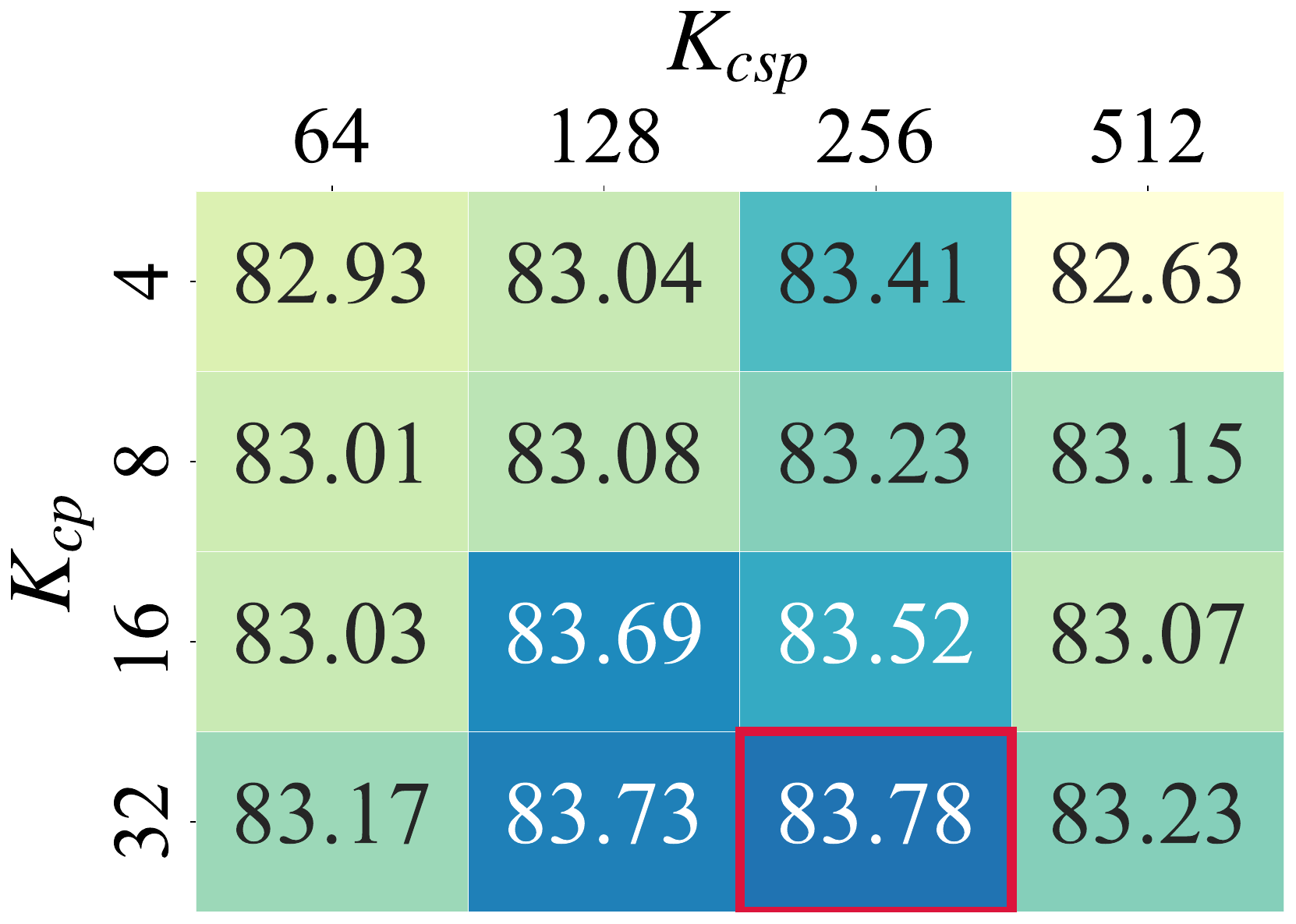}
    \label{fig:cifar10_heatmap}}
  \caption{The AUROC variation with different $K_\textrm{cp}$ and $K_\textrm{csp}$ in our VQ-Flow on MVTec AD~\cite{dataset:mvtec} and CIFAR-10~\cite{dataset:cifar10}. The best results are framed by red boxes.
  } \label{fig:heatmap}
  % \vspace{-5pt}
  \end{figure*}

\begin{table*}[t]
  \centering
    \caption{Comparison of AUROC on the semantic anomaly detection dataset CIFAR-10~\cite{dataset:cifar10} under the \textbf{Multi-Class} setting. 5 classes are assumed as normal, and the rest are considered as anomalies.}
    \label{tab:comparison-cifar10-mc}
    \renewcommand{\arraystretch}{1.2}
    \resizebox{0.8\linewidth}{!}{%
    \begin{tabular}{c|l|l|cccc|c}
    \myrule
    \multirow{2}{*}{}     & \multicolumn{1}{l|}{\multirow{2}{*}{Method}} & \multicolumn{1}{l|}{\multirow{2}{*}{Venue}} & \multicolumn{4}{c|}{Normal Indices}                                                                                   & \multirow{2}{*}{Mean} \\ \cline{4-7}
                    & \multicolumn{1}{c|}{}                        & \multicolumn{1}{c|}{}                       & \{01234\}            & \multicolumn{1}{c}{\{56789\}} & \multicolumn{1}{c}{\{02468\}} & \multicolumn{1}{c|}{\{13579\}} &                       \\ \mytinyrule  
    \multirow{5}{*}{\rotatebox{270}{\centering Separated\quad}}     & US\cite{scad:us}                                                & CVPR$^\prime$20    & 51.3          & 51.3          & 63.9          & 56.8          & 55.9          \\
                                & FCDD\cite{scad:fcdd}                                               & ICLR$^\prime$21    & 71.8          & 73.7          & 85.3          & 85.0          & 78.9          \\    
                                & PANDA\cite{scad:panda}                                           & CVPR$^\prime$21    & 66.6          & 73.2          & 77.1          & 72.9          & 72.4          \\
                                & MKD\cite{scad:mkd}                                                  & CVPR$^\prime$21    & 64.2          & 69.3          & 76.4          & 78.7          & 72.1          \\
                                & MSFlow\cite{scad:msflow}                                                    & TNNLS$^\prime$23     & 66.4        & 66.4           & 72.9         & 71.0            & 69.1               \\\mytinyrule
    \multirow{3}{*}{\rotatebox{270}{Unified\quad}}   & UniAD\cite{mcad:uniad}                                               & NeurIPS$^\prime$22 & \textbf{84.4} & 80.9          & \textbf{93.0} & 90.6          & 87.2          \\
                                & Uniformaly\cite{mcad:uniformaly}                                         & ArXiv$^\prime$23   & 79.9          & 84.7          & 90.7          & 86.4          & 85.4          \\  \cline{2-8} 
                                & \textbf{VQ-Flow}                                               & \textbf{Ours}       & 83.8          & \textbf{86.9} & 92.3          & \textbf{91.6} & \textbf{88.6} \\ \myrule
    \end{tabular}%
    }

\end{table*}

\subsection{Comparison under Multi-Class Setting}

\subsubsection{MVTec AD}

We report the detection and localization AUROC of the proposed VQ-Flow and previous separate or unified training methods on MVTec AD with multi-class setting in \Cref{tab:comparison-mvtec-sc}.
Notably, when directly adopting MSFlow~\cite{scad:msflow} to the multi-class scenario, this flow-based separated training method achieves the best performance among the separate training methods~\cite{scad:patchcore,scad:draem}, demonstrating the underlying potential for normalizing flows in MCAD.
Through the incorporation of our hierarchical vector quantization upon MSFlow, the proposed VQ-Flow is boosted to 99.5\% detection and 98.3\% localization AUROC, significantly outperforming the conventional reconstruction-based unified methods~\cite{mcad:uniad,mcad:omnial,mcad:hvqTrans}.
While the superiority of our VQ-Flow over the recent unified SOTA method Uniformaly~\cite{mcad:uniformaly} is less pronounced in performance. However, Uniformaly measures the anomaly scores by computing the distances from all clusters stored in the memory bank, leading to high computational burden ($\sim$1 FPS) and infeasible for real-time applications. In contrast, our VQ-Flow is inference-friendly, with the inference speed surpassing 30 FPS as detailed in the experimental results, satisfying the real-time requirement for industrial applications.

\subsubsection{VisA}

On the VisA dataset with complex products, the proposed VQ-Flow achieves promising AUROC performance of 95.9\% for detection and 98.9\% for localization as presented in \Cref{comparison:visa}, dramatically surpassing previous unified training methods~\cite{mcad:uniad,mcad:omnial,mcad:hvqTrans,mcad:diad}.
In particular, even compared with MSFlow~\cite{scad:msflow} trained in a single-class setting, with 95.2\%/97.8\% AUROC, our VQ-Flow still outperforms it by 0.7\% and 1.1\% in detection and localization performance.

\subsubsection{CIFAR-10}

When adapting our VQ-Flow to semantic anomaly detection, only the branch of the flow model $F_3$ with the semantic-sensitive feature map $\boldsymbol{h}^{\prime}_3$ is utilized.
With the capability of distinguishing different normal concepts, our VQ-Flow extends the superiority of MSFlow~\cite{scad:msflow} to semantic anomaly detection, and achieves the best performance among previous unified training methods~\cite{mcad:uniad,mcad:omnial,mcad:hvqTrans,mcad:diad}
As reported in \Cref{tab:comparison-cifar10-mc}, the proposed VQ-Flow significantly boosts MSFlow~\cite{scad:msflow} to achieve the best performance averaged over different normal indices, outperforming the previous methods~\cite{mcad:uniad,mcad:omnial,mcad:hvqTrans,mcad:diad} by a large margin.

\begin{table*}[t]
  \caption{Comparison of detection and localization AUROC (Det./Loc.) on MVTec AD~\cite{dataset:mvtec} under the \textbf{Single-Class} setting. }
  % The reconstruction/embedding/flow-based methods are denoted with background colors of $\textcolor{b-reconstruction}{\blacksquare}$/$\textcolor{b-embedding}{\blacksquare}$/$\textcolor{b-flow}{\blacksquare}$. \emph{Best viewed in color.}}
  \label{tab:comparison-mvtec-sc}
  \centering
  \resizebox{\textwidth}{!}{%
    \begin{tabular}{cl|>{\columncolor{reconstruction}}p{1.6cm}<{\centering}>{\columncolor{embedding}}p{1.6cm}<{\centering}>{\columncolor{flow}}p{1.6cm}<{\centering}|>{\columncolor{reconstruction}}p{1.8cm}<{\centering}>{\columncolor{reconstruction}}p{1.8cm}<{\centering}>{\columncolor{embedding}}p{1.8cm}<{\centering}>{\columncolor{flow}}p{1.8cm}<{\centering}}
  \myrule
  \multicolumn{2}{c|}{Category}                                                    & \multicolumn{3}{c|}{Separate training scheme
  }                                         & \multicolumn{4}{c}{Unified training scheme}                                                       \\ \mytinyrule
  \multicolumn{2}{c|}{Method}                                                               & SSM~\cite{tmm:uad}       & PatchCore~\cite{scad:patchcore}           & MSFlow~\cite{scad:msflow}             & UniAD~\cite{mcad:uniad}       & OmniAL\cite{mcad:omnial}          & Uniformaly~\cite{mcad:uniformaly}           & \textbf{VQ-Flow}     \\ \mytinyrule
  \multicolumn{2}{c|}{Venue}                                                          & TMM$^\prime$21     & CVPR$^\prime$22             & TNNLS$^\prime$23            & NeurIPS$^\prime$22  & CVPR$^\prime$23            & ArXiv$^\prime$23             & \textbf{Ours}        \\ \mytinyrule
  \multicolumn{1}{c|}{\multirow{10}{*}{\rotatebox{270}{Object\quad}}} & \multicolumn{1}{l|}{Bottle}      & 99.9 / 95.9                   & 100 / 98.6                      & 100 / 99.0                   & \textbf{100} / 98.1                     & 99.4 / \textbf{99.0}                 & \textbf{100} / 98.9                     & \textbf{100} / \textbf{99.0}                           \\
  \multicolumn{1}{c|}{}                         & \multicolumn{1}{l|}{Cable}      & 77.3 / 82.1                  & 99.5 / 98.4                     & 99.5 / 98.5                  & 97.6 / 96.8                    & 97.6 / 97.1                 & \textbf{100} / 97.7                     & 99.8 / \textbf{97.9}                          \\
  \multicolumn{1}{c|}{}                         & \multicolumn{1}{l|}{Capsule}     & 91.4 / 98.4                  & 98.1 / 98.8                     & 99.2 / 99.1                 & 85.3 / 97.9                    & 92.4 / 92.2                 & \textbf{98.9} / 99.0                    & 98.7 / \textbf{99.1}                         \\
  \multicolumn{1}{c|}{}                         & \multicolumn{1}{l|}{Hazelnut}     &  91.5 / 97.4                   & 100 / 98.7                      & 100 / 98.7                   & 99.9 / 94.6                    & 98.0 / 98.6                 & \textbf{100} / \textbf{99.2}                     & \textbf{100} / 99.1                           \\
  \multicolumn{1}{c|}{}                         & \multicolumn{1}{l|}{Metal Nut}    & 88.7 / 89.6                   & 100 / 98.4                      & 100 / 99.3                   & 99.0 / 95.7                    & 99.9 / 99.1                 & \textbf{100} / 97.9                     & \textbf{100} / \textbf{98.8}                           \\
  \multicolumn{1}{c|}{}                         & \multicolumn{1}{l|}{Pill}        & 89.1 / 97.8                   & 96.6 / 97.4                     & 99.6 / 98.8                  & 88.3 / 95.1                    & 97.7 / 98.6                 & 98.2 / 97.5                    & \textbf{99.3} / \textbf{98.8}                          \\
  \multicolumn{1}{c|}{}                         & \multicolumn{1}{l|}{Screw}        & 85.0 / 98.9                   & 98.1 / 99.4                     & 97.8 / 99.1                  & 91.9 / 97.4                    & 81.0 / 97.2                 & 96.2 / \textbf{99.5}                    & \textbf{97.2} / 99.1                          \\
  \multicolumn{1}{c|}{}                         & \multicolumn{1}{l|}{Toothbrush}   & 100 / 98.9                    & 100 / 98.7                      & 100 / 98.5                   & 95.0 / 97.8                    & \textbf{100 / 99.2}                  & \textbf{100} / 98.9                     & 99.5 / 98.6                          \\
  \multicolumn{1}{c|}{}                         & \multicolumn{1}{l|}{Transistor}   & 91.0 / 80.1                  & 100 / 96.3                      & 100 / 98.3                  & \textbf{100} / \textbf{98.7}                     & 93.8 / 91.7                 & 99.3 / 96.5                    &\textbf{100} / 96.9                           \\
  \multicolumn{1}{c|}{}                         & \multicolumn{1}{l|}{Zipper}       & 99.9 / 99.0                    & 99.4 / 98.8                     & 100 / 99.2                   & 96.7 / 96.0                    &\textbf{ 100 / 99.7 }                 & 99.8 / 98.5                    & 99.8 / 99.1                          \\ \mytinyrule
  \multicolumn{1}{c|}{\multirow{5}{*}{\rotatebox{270}{Texture\quad}}} & \multicolumn{1}{l|}{Carpet}       & 76.3 / 94.4                   & 98.7 / 99.0                     & 100 / 99.4                   & 99.9 / 98.0                    & 99.6 / \textbf{99.6}                 & 99.6 / 99.4                    & \textbf{100}  / 99.5                          \\
  \multicolumn{1}{c|}{}                         & \multicolumn{1}{l|}{Grid}         & 100 / 99.0                   & 98.2 / 98.7                     & 99.8 / 99.4                  & 98.5 / 98.5                    &\textbf{100}/ \textbf{99.6}                 & 98.5 / 99.4                    & 99.5 / 99.2                          \\
  \multicolumn{1}{c|}{}                         & \multicolumn{1}{l|}{Leather}      & 99.9 / 99.6                    & 100 / 99.3                      & 100 / 99.7                   & \textbf{100} / 99.5                     & 97.6 / 99.7                 & 100 / 99.6                     & \textbf{100} / \textbf{99.7}                           \\
  \multicolumn{1}{c|}{}                         & \multicolumn{1}{l|}{Tile}         & 94.4 / 90.2                   & 98.7 / 95.6                      & 100 / 98.2                   & 99.0 / 91.8                    & \textbf{100 / 99.4 }                 & 99.6 / 97.6                    & \textbf{100} / 97.9                           \\
  \multicolumn{1}{c|}{}                         & \multicolumn{1}{l|}{Wood}         & 95.9 / 86.9                   & 99.2 / 95.0                      & 100 / 97.1                   & 97.9 / 93.4                    & 98.7 / 96.9                 & 99.9 / \textbf{97.6 }                   & \textbf{100} / 96.3                           \\ \mytinyrule
  \multicolumn{2}{c|}{Average}                                                       & 92.0 / 93.9                   & 99.1 / 98.1                     & 99.7 / 98.8                 & 96.6 / 96.6                    & 97.0 / 97.8                 & 99.3 / 98.5                    & \textbf{99.6} / \textbf{98.6}                         \\ \myrule
  \end{tabular}%
  }
% \vspace{-5pt}
\end{table*}

\begin{table}[t]
  \centering
  \caption{Comparison of AUROC on the semantic anomaly detection dataset CIFAR-10~\cite{dataset:cifar10} under the \textbf{Single-Class} setting. the listed class is assumed as normal, and the rest are considered as anomalies.}
  \label{tab:comparison-cifar10-sc}
  \resizebox{\linewidth}{!}{%
  \begin{tabular}{l|
  >{\columncolor{embedding}}c 
  >{\columncolor{embedding}}c 
  >{\columncolor{embedding}}c 
  >{\columncolor{embedding}}c |
  >{\columncolor{flow}}c}
  \myrule
  Method     & US~\cite{scad:us}      & FCDD~\cite{scad:fcdd}      & MKD\cite{scad:mkd}     & UniFormaly\cite{mcad:uniformaly} & \textbf{VQ-Flow} \\ \mytinyrule
  Venue      & CVPR$^\prime$20 & ICLR$^\prime$21 & CVPR$^\prime$21 & ArXiv$^\prime$23   & \textbf{Ours}    \\ \mytinyrule
  airplane   & 78.9    & 95.0        & 90.5    & 95.7       & \textbf{97.3}    \\
  automobile & 84.9    & 96.0        & 90.4    & \textbf{97.9}       & 96.4    \\
  bird       & 73.4    & 91.0        & 79.7    & 90.6       & \textbf{95.4}    \\
  cat        & 74.8    & \textbf{90.0 }       & 77.0    & 87.1       & 88.4    \\
  deer       & 85.1    & 94.0        & 86.7    & 96.0       & \textbf{96.8}    \\
  dog        & 79.3    & \textbf{93.0}        & 91.4    & 91.0       & 92.1    \\
  frog       & 89.2    & 97.0        & 89.0    & \textbf{97.8 }      & 97.4    \\
  horse      & 83.0    & 96.0        & 86.8    & 96.6       & \textbf{98.4}    \\
  ship       & 86.2    & 97.0        & 91.5    & \textbf{97.4}       & 97.1    \\
  truck      & 84.8    & 96.0        & 88.9    & \textbf{97.3}       & 96.9    \\ \mytinyrule
  Average    & 82.0    & 95.0        & 87.2    & 94.7       & \textbf{95.6}    \\ \myrule
  \end{tabular}%
  }
  \end{table}

\subsection{Ablation Studies}

We conduct comprehensive ablation studies on MVTec AD to validate the effectiveness of each key component in our VQ-Flow, the effect of different backbone architectures, the effect of different quantization methods, and the performance within separate training schemes. Moreover, we provide the visualization of the learned prototypes in CPC and CSPC to validate the rationality of our main idea of introducing hierarchical vector quantization. 

\subsubsection{The ablation of key components in VQ-Flow}

As depicted in \Cref{ablation:components}, we first investigate the effectiveness of each key component proposed in our VQ-Flow by progressively adding them to the baseline of MSFlow~\cite{scad:msflow}, including the Conceptual Prototype Codebook (CPC), Concept-Specific Prototype Codebook (CSPC), Concept-Aware Distribution Modeling (CADM), and Positional Embedding (\texttt{PE}).
When directly performing CADM on the semantic vector $\boldsymbol{y}$ in the continuous space (id.1), the performance is significantly degraded due to the infeasibility of CADM in isolation from concept distinction.
When only incorporating the CPC (id.2) and inserting the quantized conceptual prototype $\hat{\boldsymbol{y}}$ into the flow model as the conditioning variable, the detection AUROC is improved to 98.54\%, and enabling CADM (id.4) further boosts the performance to 99.13\%.
In particular, when only CSPC is enabled (id.3), the pattern codebook degrades to Concept-Agnostic Prototype Codebook (CAPC), and the performance gains (0.16\% / 0.24\%) from id.0 to id.3 are limited compared with the synergy of CPC and CADM (id.5 v.s. id.4) with improvements of 0.38\% / 0.31\%.
The further integration of \texttt{PE} (id.6) complements the full version of our VQ-Flow, achieving the best performance of 99.52\% detection and 98.32\% localization AUROC.

\begin{table}[b]
  \caption{The effect in AUROC (Det. / Loc.) of each key component in our VQ-Flow on MVTec AD~\cite{dataset:mvtec} under \textbf{Multi-Class} setting.}
  \label{ablation:components}
  \centering
  \vspace{0.1cm}
  \resizebox{\columnwidth}{!}{
  \begin{tabular}{c|p{1.cm}<{\centering}p{0.8cm}<{\centering}p{0.8cm}<{\centering}p{0.8cm}<{\centering}|c}
  \myrule
  id. & CADM & CPC & CSPC & \texttt{PE} & Det. / Loc.            \\ \mytinyrule
  0  & \multicolumn{4}{c|}{Baseline (MSFlow~\cite{scad:msflow})}                                     & 97.52 / 97.43          \\ \mytinyrule
  1  & \multicolumn{1}{c}{\checkmark}    & \multicolumn{1}{c}{}             & \multicolumn{1}{c}{}     &    & 53.46 / 52.11          \\
  2  & \multicolumn{1}{c}{}              & \multicolumn{1}{c}{\checkmark}   & \multicolumn{1}{c}{}     &    & 98.54 / 97.82          \\
  3  & \multicolumn{1}{c}{}              & \multicolumn{1}{c}{}             & \multicolumn{1}{c}{\checkmark}    &    & 97.68 / 97.67          \\
  4  & \multicolumn{1}{c}{\checkmark}    & \multicolumn{1}{c}{\checkmark}   & \multicolumn{1}{c}{}     &    & 99.13 / 97.93          \\
  5  & \multicolumn{1}{c}{\checkmark}    & \multicolumn{1}{c}{\checkmark}   & \multicolumn{1}{c}{\checkmark}    &    & 99.51 / 98.24          \\
  6  & \multicolumn{1}{c}{\checkmark}    & \multicolumn{1}{c}{\checkmark}   & \multicolumn{1}{c}{\checkmark}    & \checkmark  & \textbf{99.52 / 98.32} \\ \myrule
  \end{tabular}%
  }
\end{table}

\subsubsection{Effect of codebook sizes in hierarchical vector quantization}

The introduction of hierarchical vector quantization in our VQ-Flow is the main factor of the success in extending normalizing flows to the multi-class anomaly detection task.
Therefore, we conduct an ablation study of varying $K_\textrm{cp}$ and $K_\textrm{csp}$ to investigate the effect of the codebook sizes in hierarchical vector quantization, visualizing the effect of the codebook sizes on MVTec AD~\cite{dataset:mvtec} and CIFAR-10~\cite{dataset:cifar10} in \Cref{fig:heatmap}.
For the industrial anomaly detection dataset MVTec AD with 15 categories, when varying $K_\textrm{cp} \in \{8, 16, 32, 64\}$ and $K_\textrm{csp} \in \{64, 128, 256, 512\}$, the AUROC of detection (\Cref{fig:heatmap} left) and localization (\Cref{fig:heatmap} middle) is marginally affected by the codebook sizes, justifying effectiveness of the hierarchical vector quantization in our VQ-Flow and the robustness to hyperparameters.
For the semantic anomaly detection dataset CIFAR-10 with 10 classes, the detection AUROC gradually climbs when increasing $K_\textrm{cp} \in \{4, 8, 16, 32\}$, and the best performance is achieved when $K_\textrm{cp}=32$ and $K_\textrm{csp}=256$ as shown in the right part of \Cref{fig:heatmap}.

\begin{figure*}[t]
  \centering
  {
    \captionsetup[subfloat]{labelsep=none,format=plain,labelformat=empty}
    \subfloat{\includegraphics[width=\textwidth]{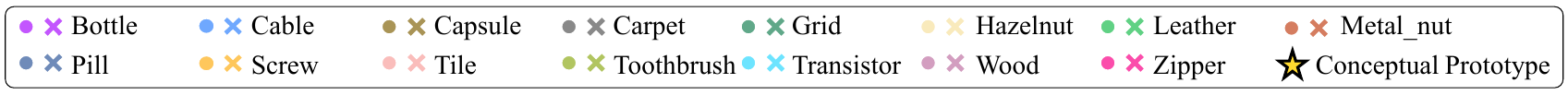}}
    \vspace{-10pt}
  }

  \setcounter{subfigure}{0}

  \subfloat[$K_\textrm{cp}$=8 on MVTec AD]{\includegraphics[width=0.3\textwidth]{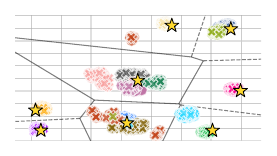}
  \label{fig:vis_mvtec_8}}
    \hfill
  \subfloat[$K_\textrm{cp}$=16 on MVTec AD]{\includegraphics[width=0.3\textwidth]{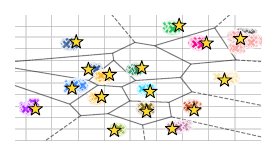}
  \label{fig:vis_mvtec_16}}
    \hfill
  \subfloat[$K_\textrm{cp}$=32 on MVTec AD]{\includegraphics[width=0.3\textwidth]{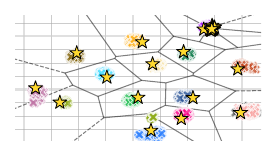}
  \label{fig:vis_mvtec_32}}
  {
    \captionsetup[subfloat]{labelsep=none,format=plain,labelformat=empty}
    \subfloat{\includegraphics[width=\textwidth]{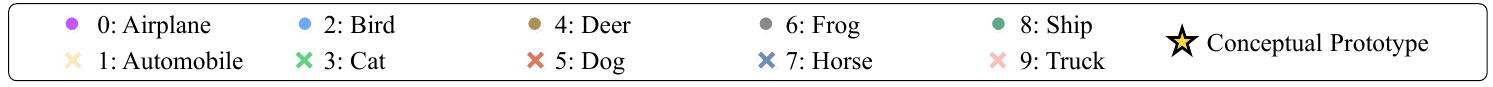}}
    \vspace{-10pt}
  }

  \setcounter{subfigure}{3}

  \subfloat[$K_\textrm{cp}$=4 on CIFAR-10]{\includegraphics[width=0.3\textwidth]{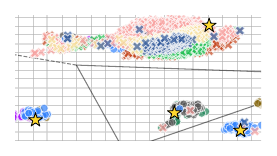}
  \label{fig:vis_cifar10_4}}
    \hfill
  \subfloat[$K_\textrm{cp}$=8 on CIFAR-10]{\includegraphics[width=0.3\textwidth]{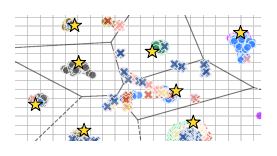}
  \label{fig:vis_cifar10_8}}
    \hfill
  \subfloat[$K_\textrm{cp}$=16 on CIFAR-10]{\includegraphics[width=0.3\textwidth]{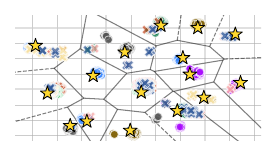}
  \label{fig:vis_cifar10_16}}
  \caption{The visualization of the conceptual prototypes with different $K_\textrm{cp}$ on MVTec AD~\cite{dataset:mvtec} and CIFAR-10~\cite{dataset:cifar10} under \textbf{Multi-Class} setting. $\bullet$ and $\boldsymbol{\times}$ denotes the \textbf{normal} or \textbf{abnormal} samples.} \label{fig:vis}
  % \vspace{-10pt}
\end{figure*}

\begin{table}[b]
  % \vspace{-10pt}
  \centering
  \caption{The effect of different backbones on the MVTec AD dataset~\cite{dataset:mvtec} under the \textbf{Multi-Class} setting, and the inference speed is measured in FPS. `WRN-50' refers to Wide ResNet-50~\cite{backbone:wide}.}
  \label{ablation:backbone}
  \resizebox{1\columnwidth}{!}{%
  \begin{tabular}{l|l|c|c|c}
  \myrule
  Method     & Backbone  & Det.    & Loc.    & FPS         \\ \mytinyrule
  UniAD\cite{mcad:uniad}      & Efficientnet~\cite{backbone:efficientnet} & 96.5          & 96.8          & 26          \\
  HVQ-Trans\cite{mcad:hvqTrans}  & Efficientnet~\cite{backbone:efficientnet} & 98.0          & 97.3          & 31          \\
  DiAD\cite{mcad:diad}       & ResNet-50~\cite{backbone:resnet}    & 97.2          & 96.8          & $\sim$1     \\
  Uniformaly\cite{mcad:uniformaly} & ViT-DINO~\cite{backbone:vit}     & 99.2          & 98.1          & $\sim$1     \\ \mytinyrule
  \textbf{VQ-Flow (Ours)}    & ResNet-18~\cite{backbone:resnet}           & 96.4          & 97.3          & \textbf{58} \\
  \textbf{VQ-Flow (Ours)}    & WRN-50~\cite{backbone:wide}         & 99.1          & 97.9          & 48          \\
  \textbf{VQ-Flow (Ours)}    & ConvNeXt~\cite{backbone:convnext}     & \textbf{99.5} & \textbf{98.3} & 33          \\ \myrule
  \end{tabular}%
  }
  % \vspace{-10pt}
\end{table}

\subsubsection{Effect of Different Backbones}

As depicted in Table~\ref{ablation:backbone}, we present a comparative analysis of the impact of different backbones within our VQ-Flow.
When progressively replacing the backbone from ResNet-18~\cite{backbone:resnet} to Wide ResNet-50 (WRN-50)~\cite{backbone:wide} and ConvNeXt-B~\cite{backbone:convnext}, the performance of our VQ-Flow is consistently improved with the increase of model capacity.
During this process, the inference speed is slightly decreased while still maintaining an acceptable speed higher than 30 FPS for real-time applications.
When compared with the previous unified methods~\cite{mcad:uniad,mcad:hvqTrans,mcad:diad,mcad:uniformaly} with diverse backbones~\cite{backbone:efficientnet, backbone:resnet,backbone:vit}, our VQ-Flow demonstrate the superiority in both performance and efficiency.

\subsubsection{Comparisons in the single class setting}
Besides recently emerging multi-class anomaly detection benchmarks, our VQ-Flow is also competitive on MVTec AD~\cite{dataset:mvtec}, VisA~\cite{dataset:visa}, and CIFAR-10~\cite{dataset:cifar10} under the single-class assumption, where the model is separately trained for each class.

As for MVTec AD~\cite{dataset:mvtec}, reported in \Cref{tab:comparison-cifar10-sc}, our VQ-Flow achieves the second-best performance in both detection and localization AUROC, outperforming all previous unified methods and only slightly inferior (0.01\% $\downarrow$) to the best-performing MSFlow~\cite{scad:msflow}.
In particular, shifting the training scheme from separated to unified, our VQ-Flow surpasses MSFlow by a large margin of 2.2\% in detection AUROC and 1.2\% in localization AUROC.
According to the comparison between \Cref{tab:comparison-mvtec-mc} and \Cref{tab:comparison-mvtec-sc}, our VQ-Flow is more competitive under the multi-class setting, which further validates our motivation of introducing hierarchical vector quantization for comprehensive normal pattern modeling.

On the semantic anomaly detection dataset CIFAR-10~\cite{dataset:cifar10}, as shown in \Cref{tab:comparison-cifar10-sc}, when only one class is assumed as normal and the rest are considered as anomalies, our VQ-Flow also surpasses the most previous methods of separate~\cite{scad:us,scad:fcdd,scad:mkd} or unified~\cite{mcad:uniformaly} training schemes, yet comparable to PANDA~\cite{scad:panda}, the SOTA method within single-class setting.
  
The superiority of our VQ-Flow under the single-class settings further demonstrates the efficiency of our motivation in utilizing hierarchical vector quantization for concept distinction and concept-specific normal pattern capturing.

\begin{figure*}[t]
  \centering
  \includegraphics[width=\textwidth]{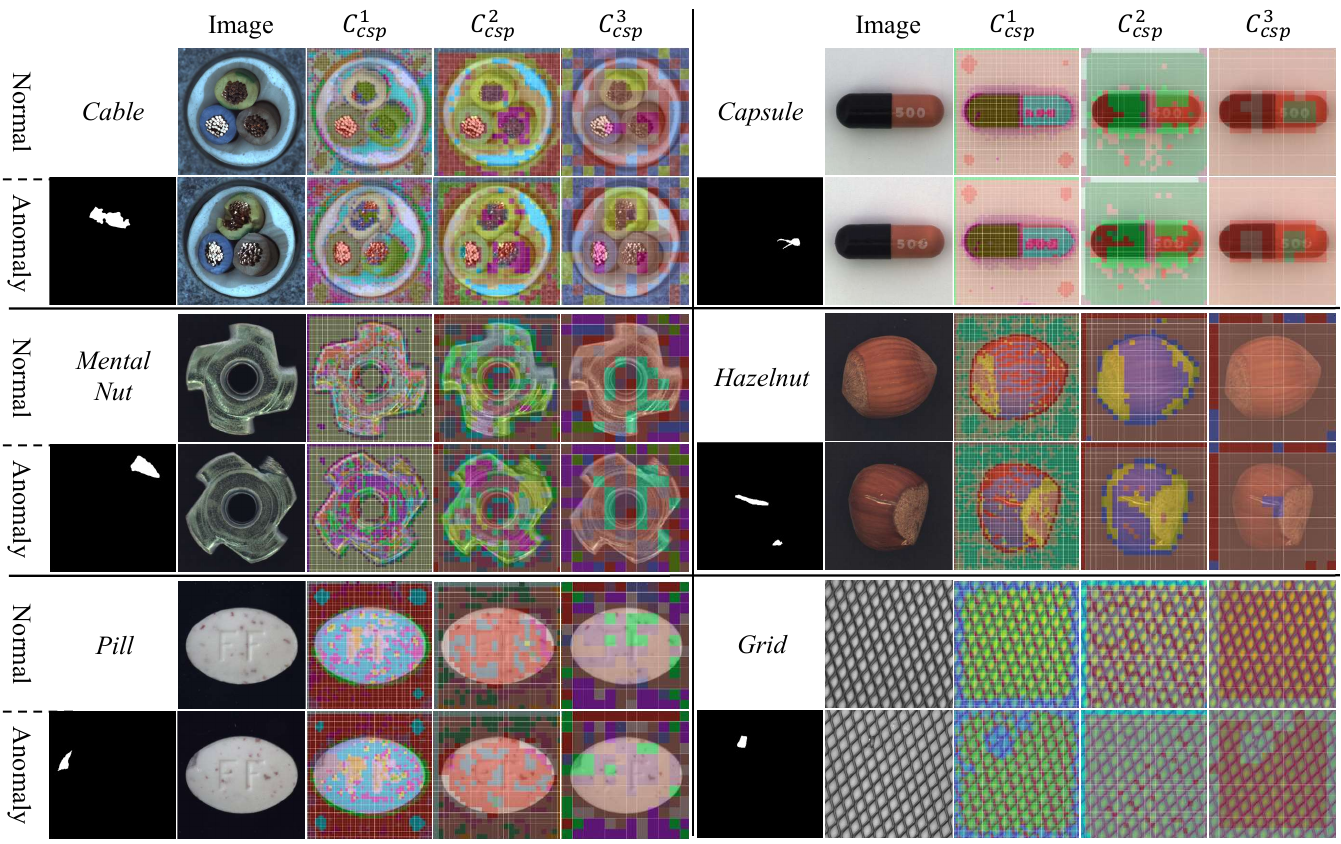}
  \caption{Intuitive visualization of the concept-specific pattern codebooks (CSPC) upon normal and anomalous samples from MVTec AD~\cite{dataset:mvtec} under the \textbf{Multi-Class} setting. All quantized feature maps ($\hat{\boldsymbol{h}^{\prime}_1}, \hat{\boldsymbol{h}^{\prime}_3}, \hat{\boldsymbol{h}^{\prime}_3}$) derived from three stages are visualized.}
  \label{fig:viz}
\end{figure*}

\subsubsection{Visualization of CPC and CSPC}

For intuitively understanding conceptual prototypes in CPC, we present the visualization of conceptual prototypes on MVTec AD~\cite{dataset:mvtec} and CIFAR-10~\cite{dataset:cifar10} in \Cref{fig:vis}. For MVTec AD with 15 categories, one conceptual prototype covers multiple categories when $K_\textrm{cp}=8$ (\Cref{fig:vis_mvtec_8}), multiple prototypes are mixed for the same category when $K_\textrm{cp}$ rises to 32 (\Cref{fig:vis_mvtec_32}).
The best visualization is achieved in \Cref{fig:vis_mvtec_16}, where $K_\textrm{cp}=16$ is the closest to the number of categories.
Particularly, in MVTec AD where the classification of anomaly is irrespective of the category, the prototypes are representative regardless of the normal or abnormal samples, which justifies the motivation of CPC for concept distinction.

Shifting to CIFAR-10 with $\{0,2,4,6,8\}$ as normal, samples from abnormal classes are pushed away from prototypes as shown in \Cref{fig:vis_cifar10_4}, where $K_\textrm{cp}=4$ is inadequate to cover 5 normal classes.
Increasing $K_\textrm{cp}$ to 8 or 16, abnormal samples are more distinguishable, situated at the boundaries between prototypes or assigned to redundant prototypes, as depicted in \Cref{fig:vis_cifar10_8} and \Cref{fig:vis_cifar10_16}.
These observations substantiate the efficacy of CPC introduced in our VQ-Flow for concept-aware anomaly detection.

We present the visualization of Concept-Specific Pattern Codebooks (CSPC) through assigning different codewords in quantized feature maps ($\hat{\boldsymbol{h}^{\prime}_1}, \hat{\boldsymbol{h}^{\prime}_3}, \hat{\boldsymbol{h}^{\prime}_3}$) with respective colors, as shown in \Cref{fig:viz}.
Consistent with the motivation of CSPC in capturing concept-specific normal patterns, the visualization of normal samples depicts the specific layout and structure of different classes of products.
In other words, the visualization of CSPC looks like the pixel art of various industrial products, whose granularity increases when moving from $\hat{\boldsymbol{h}^{\prime}_1}$ to $\hat{\boldsymbol{h}^{\prime}_3}$.
Specifically, when quantizing the feature maps derived from abnormal samples through the CSPC only trained on normal samples, the defect regions are discriminated with distinct colors, such as the examples of \emph{Hazelnut} and \emph{Grid} in \Cref{fig:viz}.

% \section{Limitation}

% Despite the superior performance of the proposed VQ-Flow over previous approaches, there is still a long way to go for the practical applications, where even a 0.1\% error rate is intolerable.
% Beyond simply classifying the normal and anomaly, anomaly reasoning identifies the underlying causes and nature of the anomalies.
% This aspect remains a challenging open problem within the anomaly detection domain and has not yet been addressed by our VQ-Flow.

\section{Conclusion}

We are the first to extend normalizing flows to the multi-class anomaly detection task through the integration of hierarchical vector quantization, proposing VQ-Flow. 
By capturing concept-specific normal patterns and modeling the intricate multi-class data distribution, VQ-Flow is capable of distinguishing anomalies from normal samples within a multi-class setting.
The concept-aware conditional distribution modeling further enables VQ-Flow to faithfully model the intricate multi-class data distribution.
Superior performance in both industrial and semantic anomaly detection tasks and comprehensive ablation studies demonstrate the effectiveness of the proposed VQ-Flow towards multi-class setting, highlighting the potential of normalizing flows in anomaly detection.
The real-time inference speed of the proposed VQ-Flow is also competitive, making it practical for real-world applications.

\FloatBarrier

\bibliographystyle{IEEEtran}
\bibliography{main}

\end{document}